\documentclass[journal]{new-aiaa} 
\usepackage[utf8]{inputenc}

\usepackage{graphicx}
\usepackage{amsmath}
\usepackage[bf]{subfigure}
\usepackage[version=4]{mhchem}
\usepackage{siunitx}
\usepackage{longtable,tabularx}
\title{Error-Covariance Analysis of Monocular Pose Estimation Using Total Least Squares}

\author{Saeed Maleki\footnote{Ph.D.~Candidate, Department of Mechanical and Aerospace Engineering, Email: saeedmal@buffalo.edu}
}\author{John L.~Crassidis\footnote{SUNY Distinguished Professor, Moog Endowed Chaired Professor of Innovation, Department of Mechanical and Aerospace Engineering, Email: johnc@buffalo.edu. Fellow AIAA.}}
\affil{University at Buffalo, State University of New York, Amherst, NY, 14260-4400}

\author{Yang Cheng\footnote{Associate Professor, Department of Aerospace Engineering, Email: cheng@ae.msstate.edu.  Associate Fellow AIAA.}}
\affil{Mississippi State University, Mississippi State, MS, 39762}
\author{Matthias Schmid\footnote{Research Assistant Professor, Department of Automotive Engineering, Email: schmidm@clemson.edu.}}
\affil{Clemson University, Clemson, SC, 29634}

\begin{document}

\maketitle

\begin{abstract}
This study presents a theoretical structure for the monocular pose estimation problem using the total least squares. The unit-vector line-of-sight observations of the features are extracted from the monocular camera images. First, the optimization framework is formulated for the pose estimation problem with observation vectors extracted from unit vectors from the camera center-of-projection, pointing towards the image features. The attitude and position solutions obtained via the derived optimization framework are proven to reach the Cram\'er-Rao lower bound under the small angle approximation of the attitude errors. Specifically, The Fisher Information Matrix and the Cram\'er-Rao bounds are evaluated and compared to the analytical derivations of the error-covariance expressions to rigorously prove the optimality of the estimates.
The sensor data for the measurement model is provided through a series of vector observations, and two fully populated noise-covariance matrices are assumed for the body and reference observation data. The inverse of the former matrices appear in terms of a series of weight matrices in the cost function.  The proposed solution is simulated in a Monte-Carlo framework with 10,000 samples to validate the error-covariance analysis.

\end{abstract}

\section{Introduction}\label{sec:introduction}
The simultaneous localization and mapping (SLAM) problem focuses on whether a mobile robot is able to assess situational awareness when put in an unknown environment and simultaneously make a map of the obstacles, as well as the possible landmarks on the surrounding objects.

In the SLAM problem, the localization part that determines the orientation (attitude) of the vehicle can be stated as an attitude estimation problem. The attitude estimation-only problem can be classified into two main categories \cite{markley2014fundamentals}.
The first category determines the attitude with respect to a reference frame using the vector observation data from the surrounding landmark features in a deterministic manner using multiple simultaneous vector-observations.  For observability purposes, there should be at least two non-collinear features detected with respect to the body frame of the vehicle.

The second category focuses on filtering the observation data from the environment by fusing them as a measurement model with a dynamic motion model of the vehicle trajectories in terms of a state estimation problem \cite{survey_nonlin_att}.
Methods based on the first category can be used to mitigate the transients and divergence of the filtering approaches, while the second can provide the estimation even with one observation vector \cite{psiaki1990tam}.  Every attitude determination problem involves finding the extremum of a cost function.

One of the early cost functions introduced for the attitude determination is the Wahba's problem \cite{wahba1965least}. Many solutions to solve this problem have been shown since its introduction \cite{markley2000quaternion}.
Wahba's problem is also related to the attitude and position determination problem \cite{eggert1997estimating}, which is called pose estimation. This current work addresses the pose estimation problem in a monocular vision setup.

Pose estimation is a crucial part of the vision-based navigation applications, because an accurate pose estimate typically needs to be leveraged for the vehicle controller to avoid undesirable efforts. Pose estimation solutions using imagery data-sets can be classified in two main approaches \cite{kelsey2006vision}: model-based and non-model-based approaches.

Model-based solutions use a presumed model of the object using their dimensions, shape or texture, and corresponds the features to solve for the model parameters.
Template matching \cite{jurie2002real} and contour tracking \cite{isard1998condensation} are examples for this class of pose estimation solutions.
Non-model-based approaches do not leverage an object model, but rather determine the pose from a sequence of images such as structure from motion problem \cite{schonberger2016structure}.
One solution is information fusion of high-frequency inertial sensors and a relatively low-frequency monocular camera for the pose estimation of a rigid body \cite{rehbinder2003pose}.
Reference \cite{sharma2016comparative} provides a comparative study of different methods of initial pose estimation from monocular camera data for a spacecraft mission application.
Attitude estimation from the two rotated unit-vectors using the geometric algebra is an interesting rigorous mathematical approach and has been studied in \cite{lasenby1998new}.
This solution can be used for the monocular vision applications as well.
In \cite{huang2002motion}, a review of different pose estimation methods is provided using monocular or stereo camera image data-sets, in which the available features are either the coordinates of 3D points in the world or their corresponding 2D projections on the image plane.
Monocular vision can be used for pose estimation of the UAV's by only using an off-board camera \cite{fu2017robust}.
ORB-SLAM \cite{mur2015orb} is one of the real-time solutions for the SLAM problem using monocular camera which is robust to clutter.
Deep learning-based approaches also provide a series of solutions for relative pose estimation in vision applications \cite{melekhov2017relative, li2018deepim, toshev2014deeppose}.
The recovery of feature depth as well as the attitude matrix is given in an iterative approach in \cite{zhang2010monocular}.
The pose determination problem is solved in \cite{dornaika1999pose} using point and line correspondences.
The first is based on the weak perspective model of the monocular camera, and the second is a first-order approximation of the perspective model. The line correspondence solution has better convergence properties and more accurate estimates compared to the point correspondence, but comes at the price of higher computational complexity \cite{rehbinder2003pose}. Structure from motion is one of the super-sets of pose estimation problems, in which the 3D coordinates of features as well as the relative pose of the camera with respect to a reference frame are estimated.
For an in-depth review of the structure from motion algorithms and solutions, the reader is referred to \cite{schonberger2016structure}.
Vision-based pose estimation has many applications in human-computer interaction.
Monocular or multiple cameras can be used to assess the orientation of hand pose \cite{ErolAli2007Vhpe}, in which several degrees of freedom are solved as unknowns.

While working with visual measurement data, the objects of interest first need to be segmented. Features are usually then detected, which are finally corresponded to each other in different image scans.
Image segmentation is a fundamental problem in SLAM using visual data. This has many applications in robotics \cite{forsyth2012computer} such as the autonomous driving, in which the surface, pedestrians, cars and bikes, etc., need to be detected. 
Image segmentation can be stated as classifying the pixels with different labels (semantic segmentation) or separating the objects of interest as a whole (instance segmentation).
There are a variety of image segmentation methods that range from  simple approaches, such as like kmeans \cite{dhanachandra2015image}, or more computationally involved and powerful methods, such as deep-learning based methods \cite{chen2017rethinking} or Markov random fields \cite{plath2009multi}.
A survey of  deep-learning approaches to solve the image segmentation problem is shown in \cite{minaee2021image,chouhan2018soft,zaitoun2015survey}. Before initiating the actually pose determination process, first the feature extraction and feature correspondence problems must be solved.
Features are intuitively the important points in an image that can be found in other images of the same scene in a unique way.
Feature extraction techniques are categorized into low-level and high-level approaches.
Low-level methods find the local features in an image.
The image instances can be derived from many approaches, such as the Canny edge detector \cite{canny1986computational} to SIFT \cite{lowe2004distinctive} and SURF \cite{bay2006surf}.
High-level features are more focused on detection of more complex features, such as lines and circles and other generic shapes. The Hough transform \cite{hough1962method} and generalized Hough transform \cite{ballard1981generalizing} are some basic solutions using high-level features. A more elaborate review of feature extraction methods is provided in \cite{salahat2017recent}.
Feature correspondence is also one of the basic problems in image processing and has many applications in visual SLAM \cite{davison2007monoslam}.
Grid-based motion statistics is one of the recent computationally efficient methods for feature matching \cite{Bian_2017_CVPR}.
A comparative study of the 2D feature matching methods is provided in \cite{zhao2019comparative}.

The assumptions of this work are that the segmentation, feature extraction and feature correspondence steps are solved.

The main focus then will be to solve for the pose estimation problem by utilizing the pairs of matched features in the format of vector observation pairs. The pose estimation from vector observation pairs is well-established in the literature \cite{hashim2020attitude}.
But the statistical analysis of the estimates, in an optimal way that achieves the Cram\'er-Rao lower bound (CRLB) \cite{alma9938879548204803}, has not been studied well.
The error-covariance expressions in Lidar data sets with the most generic positive definite matrices for the vector-measurement covariance are derived in \cite{maleki2021total} and \cite{maleki2022thesis}.
The statement of the pose estimation problem as a total least squares (TLS) problem has been elaborated in \cite{cheng2021optimal}, in which there are errors in both the ``design matrix'' and the  measurement observation \cite{crassidis2019maximum}.
In this paper, the problem of the most general case of sensor uncertainty in which there are correlations between the different features in the reference observation vectors, as well as the body observation vectors, is solved for the unit observation vectors from the camera pointing towards the features.
Furthermore, error-covariance expressions are derived that achieve the CRLB.

\section{Overview of Linear Least Squares and Total Least Squares}\label{intro_tls}
This section provides a brief introduction to linear and total least squares, and how they are related and their differences.
For a more in-depth review of the TLS, see \cite{golub1973some, markovsky2007overview, golub1980analysis}. 

Consider the measurement model of the form 
\begin{equation}\label{eqn:meas_model_LS}
    \tilde{\mathbf{y}}=H\mathbf{x}+\Delta\mathbf{y}
\end{equation}
where $H $ is a $m\times n$ deterministic matrix with no errors, $\mathbf{x}$ is the $n\times 1$ vector of unknowns, $\tilde{\mathbf{y}}$ is the $m \times 1$ measurement vector, and $\Delta\mathbf{y}$ is the $m \times 1$ measurement error-vector.
The least squares estimate of $\mathbf{x}$ is given by solving the following problem:
\begin{equation}\label{cost_ls}
\begin{gathered}
    \underset{\mathbf{\hat{x}}}{\min}\text{\ }J=\frac{1}{2}\Delta\mathbf{y}^T\Delta\mathbf{y} \\
    \text{subject to: }\mathbf{\hat{y}}=H\mathbf{\hat{x}}
\end{gathered}
\end{equation}
where the number of measurement samples stacked vertically in the vector $\tilde{\mathbf{y}}$ should be more than the number of unknowns, and $H$ should have at least rank $n$, for the problem to be observable.
The main underlying assumption in the statistical analysis of least squares is that $\tilde{\mathbf{y}}$ has a Gaussian distribution with the conditional likelihood function given by
\begin{equation}\label{eqn:liklihood_ls}
    p(\mathbf{\tilde{y}}|\mathbf{x})=\frac{1}{(2\pi)^{\frac{m}{2}}\big[\det(R_{yy})\big]^{\frac{1}{2}}}\text{exp}\left\{-\frac{1}{2}(\mathbf{\tilde{y}}-H \mathbf{x})^TR^{-1}_{yy}(\mathbf{\tilde{y}}-H \mathbf{x})\right\}
\end{equation}
where the distribution mean is denoted by $H\mathbf{x}$ and the covariance is $R_{yy}$. 
Because of the properties of the exponential function, maximizing the likelihood function \ref{eqn:liklihood_ls} is equivalent to minimizing the negative of the log-likelihood. 
The mean and error-covariance of the estimate are given by
\begin{subequations}
\begin{gather}
    \text{E}\{\mathbf{\hat{x}}\}=\mathbf{x}\\
    \text{cov}\{\mathbf{\hat{x}}\}=H^T R^{-1}_{yy} H
\end{gather}
\end{subequations}
which shows that the least squares estimate is unbiased.

As stated previously,the design matrix $H$ in the least squares measurement model in Eq.~\eqref{eqn:meas_model_LS} has no errors.
If this underlying assumption does not exist anymore, which happens in many applications, as will be seen in the SLAM problem in Section \ref{tls_derivation}, then another formulation must be used to consider the errors in the design matrix, which leads to the TLS problem, with paramaters defined by

\begin{subequations}
\begin{gather}
  \tilde{\mathbf y}=\mathbf{y}+\Delta\mathbf{y}\\
  \mathbf{y}=H\mathbf{x}\\
  \tilde{H}=H+\Delta{H}
\end{gather}
\end{subequations}
where  $\Delta H$ shows the errors in the design matrix. Consider the following augmented matrix:
\begin{equation}
    D=\begin{bmatrix}H & \mathbf{y}\end{bmatrix}
\end{equation}
The conditional likelihood function of the TLS problem is defined by 
\begin{equation}
     p(\tilde{D}|D)=\frac{1}{(2\pi)^{\frac{m}{2}}\big[\det(R)\big]^{\frac{1}{2}}}\text{exp}\left\{-\frac{1}{2}\text{vec}(\tilde{D}-D)^TR^{-1}\text{vec}(\tilde{D}-D)\right\}
\end{equation}
where vec operator stacks all columns of a matrix in a single column.
The maximum likelihood approach for this cost function leads to the minimization of the log-likelihood function as
\begin{equation}\label{eqn:cost_TLS_D}
\begin{gathered}
    J(\hat{D})=\frac{1}{2}\text{vec}(\tilde{D}-\hat{D})^T R^{-1} \text{vec}(\tilde{D}-\hat{D})\\
    \text{subject to}: \hat{D}\,{\hat{\mathbf z}}=\textbf{0} \\
\end{gathered}
\end{equation}
where $\hat{\mathbf{z}}=\begin{bmatrix}\hat{\mathbf{x}} & -1\end{bmatrix}^T$. A unique solution for this problem can be obtained if $\text{rank}(D)=n$.
Also, $R$ is the covariance matrix that accounts for the errors in both $\tilde{\mathbf y}$ and $\tilde{H}$.
Although the TLS solution is known to be biased, the TLS problem is proven to reach the CRLB \cite{cramer1999mathematical} for the estimate error-covariance to within first-order error-terms \cite{crassidis2019maximum}.
Closed-form solutions for the TLS problem are possible only when $R$ is an isotropic matrix.
\section{Total Least Squares Derivation for Pose Determination}\label{tls_derivation}
\begin{figure}
  \centering
  \includegraphics[width=4in]{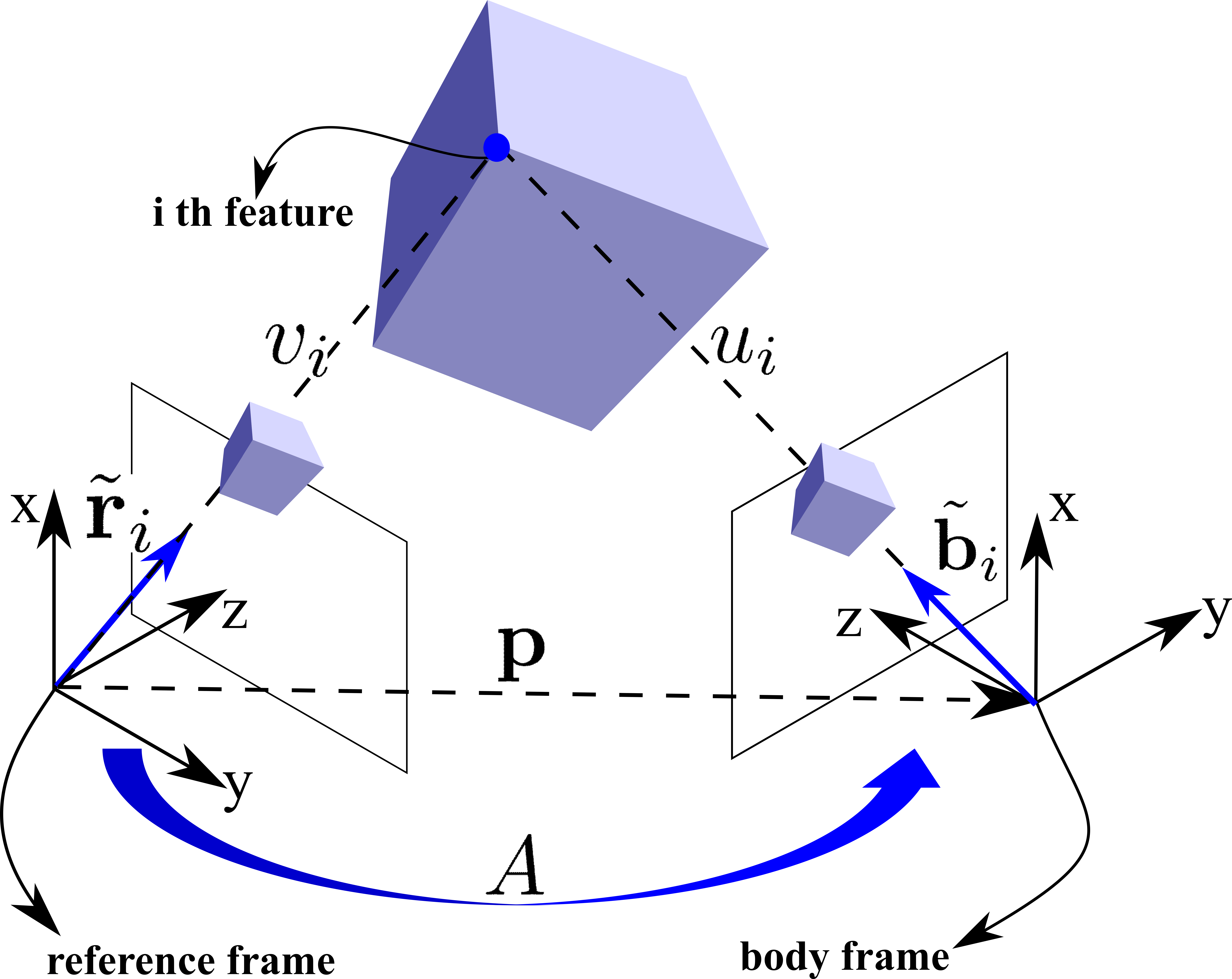}\\
  \caption{Geometric interpretation of the monocular pose estimation problem.}\label{fig:geometry_calibration}
\end{figure}

The schematic relating the pose of the reference frame to the body frame is shown in Figure \ref{fig:geometry_calibration}.
There are two components that need to be estimated: 1) a translation vector $\mathbf{p}$ that connects the center of the reference and body frames, and 2) an attitude matrix $A$ for the relative orientation of the unit vectors of the coordinate systems.
As mentioned in section \ref{sec:introduction}, the sphere image surface instead of conventional image plane is utilized in this paper. Therefore, a series of line-of-sight measurements in the form of unit vectors is used.
The terms $u_i$ and $v_i$ denote the depth of the $i^{th}$ feature on the object of interest.
The $\tilde{\mathbf{b}}_i$ and $\tilde{\mathbf{r}}_i$ denote the unit vector directions of the line-of-sight for feature $i$ in reference and body frames, respectively.
The constraint that relates the unit vectors of projection from both coordinate systems is given by
\begin{align}
    u_i\mathbf{b}_i=v_i A \mathbf{r}_i-\mathbf{p}\text{,\ \ \ }i=1,...,n\label{eqn:perfect_constraint_sev}
\end{align}
In which $n$ shows the number of features in every scan of the monocular camera.
In this study, the camera is assumed to be intrinsically calibrated and only the extrinsic parameters of attitude and position are sought.
Equation~\eqref{eqn:perfect_constraint_sev} is a measurement model for the estimation problem.
In this paper, the $\tilde{.}$ shows the measurement signal and the $\hat{.}$ refers to the estimated value of the unknowns.
The Eq.~\eqref{eqn:perfect_constraint_sev} can be written as
\begin{equation}
\begin{split}
    u_i\mathbf{b}_i&=v_i A\mathbf{r}_i-\mathbf{p} \\
    &=v_i \begin{bmatrix}\mathbf{r}^T_i & \mathbf{0}_{1\times3} & \mathbf{0}_{1\times3}\\
    \mathbf{0}_{1\times3} & \mathbf{r}^T_i & \mathbf{0}_{1\times3} \\
    \mathbf{0}_{1\times3} & \mathbf{0}_{1\times3} & \mathbf{r}^T_i  \end{bmatrix}\begin{bmatrix}\mathbf{a}_1\\
    \mathbf{a}_2\\
    \mathbf{a}_3\end{bmatrix}-\mathbf{p}\\
    &=u_i H_i\mathbf{x}-\mathbf{p} \\ \label{eqn:det_tls_sensor_model}
    &\equiv\mathbf{y}_i
\end{split}
\end{equation}
where $H_i$ is the design matrix for the $i^{th}$ feature and $\mathbf{a}_i$, $i=1,\,2,\,3$, are the columns of the attitude matrix $A$.
The perfect measurement model however is not realistic because of noise in the design matrix as well as the observation vectors of Eq.~\eqref{eqn:det_tls_sensor_model}.
Therefore, in the actual version of the sensor model in Eq.~\eqref{eqn:det_tls_sensor_model}, the following relation is used:
\begin{equation}
\begin{split}
    \left(\tilde{u}_i-\Delta u_i\right)\left({\tilde{\mathbf b}_i}-\boldsymbol{\Delta}\mathbf{b}_i\right)&=\left(\tilde{v}_i-\Delta v_i\right)A({\tilde{\mathbf r}_i}-\boldsymbol{\Delta}\mathbf{r}_i)-\mathbf{p}\\
    &=(\tilde{H}_i-\Delta H_i){\mathbf x}-\mathbf p \\
    &\equiv{\tilde{\mathbf y}_i}-\boldsymbol{\Delta} \mathbf{y}_i
\end{split}
\end{equation}
where the design matrix $\tilde{H}_i$ and the observation vector $\tilde{\mathbf{y}}_i$ have the errors of $\Delta H_i$ and $\Delta \mathbf{y}_i$, respectively.
Because the model is linear in terms of the unknowns $\mathbf{x}$ and the translation vector $\mathbf{p}$, then the problem can be posed using a TLS formulation with the constraint 
\begin{equation}
 \hat{u}_i{\hat{\mathbf b}_i}=\hat{v}_i\hat{A}{\hat{\mathbf r}_i}-{\hat{\mathbf p}}
\end{equation}
which shows the structure of a TLS problem because of the erroneous design matrix.
It is already well-known that an image from a monocular camera does not provide us with depth information of the features.
To avoid observability problems, suppose that rough measurements with large covariance of all the scale factors in the both frames are provided.
Therefore, assume that $u_i$ and $v_i$ are measured and denoted by $\tilde{u}_i$ and $\tilde{v}_i$ with the covariances of $R_{u_i}$ and $R_{v_i}$.
The cost function is
\begin{equation}
    \begin{split}
        J(\tilde{u}_i,\hat{u}_i,\tilde{v}_i,\hat{v}_i,\tilde{\mathbf{b}}_i,\hat{\mathbf{b}}_i,\tilde{\mathbf{r}}_i,\hat{\mathbf{r}}_i)&=\frac{1}{2}\sum_{i=1}^n (\tilde{u}_i-\hat{u}_i)^T R^{-1}_{u_i}(\tilde{u}_i-\hat{u}_i)+\frac{1}{2}\sum_{i=1}^n (\tilde{v}_i-\hat{v}_i)^T R^{-1}_{v_i}(\tilde{v}_i-\hat{v}_i)\\
        &+\frac{1}{2}\sum_{i=1}^n (\tilde{\mathbf{b}}_i-\mathbf{\hat{b}}_i)^TR_{b_i}^{-1}(\tilde{\mathbf{b}}_i-\mathbf{\hat{b}}_i)+(\tilde{\mathbf{r}}_i-\mathbf{\hat{r}}_i)^TR_{r_i}^{-1}(\tilde{\mathbf{r}}_i-\mathbf{\hat{r}}_i)\label{eqn:cost_sf_1_sev}
    \end{split}
\end{equation}
subject to the constraints
\begin{subequations}
\begin{gather}
    \hat{u}_i\mathbf{\hat{b}}_i=\hat{v}_i\hat{A}\mathbf{\hat{r}}_i-\mathbf{\hat{p}}\text{\ \ \ }i=1,...,n\label{lin_const_sev}\\
    \hat{A}\hat{A}^T=I_{3\times3},\text{\ } \text{det}(\hat{A})=1
\end{gather}
\end{subequations}
Define the vector $\mathbf{d}_i$ as follows:
\begin{equation}
    \mathbf{d}_i=\begin{bmatrix}\mathbf{r}_i\\\mathbf{b}_i\end{bmatrix}
\end{equation}
The cost function needs to augmented to include the constraint in Eq.~\eqref{lin_const_sev}, which is accomplished using Lagrange multiples, denoted by $\boldsymbol{\lambda}_i$:
\begin{equation}
    \begin{split}
        J_a&=\frac{1}{2}\sum_{i=1}^n (\tilde{u}_i-\hat{u}_i)^T R^{-1}_{u_i}(\tilde{u}_i-\hat{u}_i)+\frac{1}{2}\sum_{i=1}^n (\tilde{v}_i-\hat{v}_i)^T R^{-1}_{v_i}(\tilde{v}_i-\hat{v}_i)\\
        &+\frac{1}{2}\sum_{i=1}^n (\tilde{\mathbf{d}}_i-\mathbf{\hat{d}}_i)^TR_{i}^{-1}(\tilde{\mathbf{d}}_i-\mathbf{\hat{d}}_i)+2\boldsymbol{\lambda}_i^T(-\hat{u}_i\mathbf{\hat{b}}_i+\hat{v}_i\hat{A}\mathbf{\hat{r}}_i-\mathbf{\hat{p}})\label{eqn:cost_init_sev}
    \end{split}
\end{equation}
The necessary condition of this augmented cost function for $\mathbf{\hat{d}}_i$ will be
\begin{equation}
    \frac{\partial J_a}{\partial \mathbf{\hat{d}}_i}=-R^{-1}_i(\mathbf{\tilde{d}}_i-\mathbf{\hat{d}}_i)+\hat{S}^T_i\boldsymbol{\lambda}_i=\mathbf{0}\label{eqn:delJdi_sev}
\end{equation}
in which
\begin{subequations}
\begin{gather}
    R_i\equiv \text{E}\{\boldsymbol{\Delta}\mathbf{d}_i\boldsymbol{\Delta}\mathbf{d}^T_i\}=\begin{bmatrix}R_{r_i} & 0_{3\times3}\\
    0_{3\times3}& R_{b_i}\end{bmatrix}\label{eq:P_Ddi}\\
    \hat{S}_i=\begin{bmatrix}\hat{v}_i\hat{A}&-\hat{u}_i I_{3\times3}\end{bmatrix}\\
    R_{r_i}=\text{E}\{\boldsymbol{\Delta}\mathbf{r}_i\boldsymbol{\Delta}\mathbf{r}^T_i\}\\
    R_{b_i}=\text{E}\{\boldsymbol{\Delta}\mathbf{b}_i\boldsymbol{\Delta}\mathbf{b}^T_i\}\\
    R_{u_i}=\text{E}\{\Delta u^2_i\}\\
    R_{v_i}=\text{E}\{\Delta v^2_i\}
\end{gather}
\end{subequations}
From the previous necessary condition in Eq.~\eqref{eqn:delJdi_sev}, for the features in an image scan $i=1,...,n$, the following is given:
\begin{equation}
    \mathbf{\hat{d}}_i=\mathbf{\tilde{d}}_i-R_i\hat{S}^T_i\boldsymbol{\lambda}_i\label{eqn:def_dhati_sev}
\end{equation}
Then the Lagrangian multiplier $\boldsymbol{\lambda}_i$ can be computed as
\begin{equation}
    \boldsymbol{\lambda}_i=Q^{-1}_{\hat{\lambda}_i}(\hat{S}_i\mathbf{\tilde{d}}_i-\mathbf{\hat{p}})\label{eqn:def_lami_sev}
\end{equation}
in which 
\begin{equation}
    Q_{\hat{\lambda}_i}=\hat{S}_i R_i\hat{S}^T_i
\end{equation}
The cost function in Eq.~\eqref{eqn:cost_init_sev} can be re-written as
\begin{equation}
    \begin{split}
        J_a&=\frac{1}{2}\sum_{i=1}^n (\tilde{u}_i-\hat{u}_i)^T R^{-1}_{u_i}(\tilde{u}_i-\hat{u}_i)+\frac{1}{2}\sum_{i=1}^n (\tilde{v}_i-\hat{v}_i)^T R^{-1}_{v_i}(\tilde{v}_i-\hat{v}_i)\\
        &+\frac{1}{2}\sum_{i=1}^n (\tilde{\mathbf{d}}_i-\mathbf{\hat{d}}_i)^T R^{-1}_i(\tilde{\mathbf{d}}_i-\mathbf{\hat{d}}_i)+2\boldsymbol{\lambda}^T_i(\hat{S}_i\mathbf{\hat{d}}_i-\mathbf{\hat{p}})
    \end{split}
\end{equation}
From Eq.~\eqref{eqn:def_dhati_sev} and Eq.~\eqref{eqn:def_lami_sev}
\begin{equation}
    \mathbf{\tilde{d}}_i-\mathbf{\hat{d}}_i=R_i \hat{S}^T_i Q^{-1}_{\hat{\lambda}_i}(-\hat{u}_i\mathbf{\tilde{b}}_i+\hat{v}_i\hat{A}\mathbf{\tilde{r}}_i-\mathbf{\hat{p}})\label{eqn:def_meas_res}
\end{equation}
Then the cost function in Eq.~\eqref{eqn:cost_sf_1_sev} can be rewritten as
\begin{equation}
    \begin{split}
        J&=\frac{1}{2}\sum_{i=1}^n (\tilde{u}_i-\hat{u}_i)^T R^{-1}_{u_i}(\tilde{u}_i-\hat{u}_i)+\frac{1}{2}\sum_{i=1}^n (\tilde{v}_i-\hat{v}_i)^T R^{-1}_{v_i}(\tilde{v}_i-\hat{v}_i)\\
    &+\frac{1}{2}\sum_{i=1}^n \left(\hat{u}_i\mathbf{\tilde{b}}_i-\hat{v}_i\hat{A}\mathbf{\tilde{r}}_i+\mathbf{\hat{p}}\right)^T Q^{-1}_{\hat{\lambda}_i} \left(\hat{u}_i\mathbf{\tilde{b}}_i-\hat{v}_i\hat{A}\mathbf{\tilde{r}}_i+\mathbf{\hat{p}}\right)\label{eqn:cost_sf_1_sev_final}
    \end{split}
\end{equation}
It is to be noted that the TLS solution has an extra capability of providing estimates for the vector observations, which are the positions of the vehicle with respect to the landmark features in the environment, as well the unknown pose.
This can be compared to the other solutions for the pose estimation problem in a sense that makes the TLS a proper solution for the more generic SLAM problem.
Also the TLS solution provides the measurement residuals based on Eq.~\eqref{eqn:def_meas_res}, which are the difference of the measurements with the estimates.
This is very useful to provide a measure of the accuracy of the algorithms, as well as a good reference for the tuning of the measurement covariances for the real applications.
\subsection{Linear Attitude Measurement model}
The relation between the true and estimated attitude matrix can be expressed as
\begin{equation}
    \hat{A}=\text{exp}(-[\boldsymbol{\delta}\boldsymbol{\alpha}\times])A
\end{equation}
where $[.\times]$ denotes the cross product matrix of a vector \cite{markley2014fundamentals}.
Using a small angle assumption which is a first-order approximation of the attitude error, the attitude estimate can be written as
\begin{equation}
    \hat{A}\approx\big(I_{3\times3}-[\boldsymbol{\delta}\boldsymbol{\alpha}\times ]\big)A\label{eqn:att_error_1st}
\end{equation}
Note that for a first-order approximation of the covariance, a second-order approximation of the cost function is required.
The error terms inside of the summations in Eq.~\eqref{eqn:cost_sf_1_sev_final} is approximated as
\begin{equation}\label{eq:first_order_sev}
    \begin{split}
        \hat{u}_i\mathbf{\tilde{b}}_i -\hat{v}_i\hat{A}\mathbf{\tilde{r}}_i+\mathbf{\hat{p}}&=(u_i+\delta u_i)(\mathbf{b}_i+\boldsymbol{\Delta}\mathbf{b}_i)-(v_i + \delta v_i)\big(I_{3\times3}-[\boldsymbol{\delta} \boldsymbol{\alpha}\times ]\big)A(\mathbf{r}_i+\boldsymbol{\Delta}\mathbf{r}_i)+(\mathbf{p}+\boldsymbol{\delta}\mathbf{p})\\
    &\approx(u_i\mathbf{b}_i-v_i A \mathbf{r}_i+\mathbf{p})-G_i \boldsymbol{\delta}\mathbf{x} +\boldsymbol{\Delta} \mathbf{a}_i
    \end{split}
\end{equation}
where the first term is zero based on the constraint in Eq.~\eqref{eqn:perfect_constraint_sev} and
\begin{subequations}
    \begin{gather}
        G_i=\begin{bmatrix}v_i[A\mathbf{r}_i\times] & -I_{3\times3} & 0_{3\times2(i-1)} & -\mathbf{b}_i & A\mathbf{r}_i & 0_{3\times2(n-i)}\end{bmatrix}_{3\times(6+2n)}\\
    \boldsymbol{\Delta}\mathbf{a}_i=u_i\boldsymbol{\Delta}\mathbf{b}_i-v_i A\boldsymbol{\Delta}\mathbf{r}_i\label{eq:def_delta_ai_sev}\\
    \boldsymbol{\delta}\mathbf{x}=\begin{bmatrix}\boldsymbol{\delta}\boldsymbol{\alpha}^T & \boldsymbol{\delta}\mathbf{p}^T & \delta u_1 & \delta v_1 & ... & \delta u_i & \delta v_i & ... & \delta u_n & \delta v_n\end{bmatrix}^T_{(6+2n)\times 1}
    \end{gather}
\end{subequations}
It is to be noted that the term $Q_{\hat{\lambda}_i}$ in Eq.~\eqref{eqn:cost_sf_1_sev_final} is also a function of the unknowns and has an approximation within first order of estimate errors $\boldsymbol{\delta}\mathbf{x}$ which is derived from
\begin{equation}
    \begin{split}
        Q_{\hat{\lambda}_i}=&\hat{v}^2_i\hat{A}R_{r_i}\hat{A}^T+\hat{u}^2_i R_{b_i}\\
    =&(v_i+\delta v_i)^2\big(I_{3\times3}-[\boldsymbol{\delta} \boldsymbol{\alpha}\times]\big)A R_{r_i}  A^T\big(I_{3\times3}-[\boldsymbol{\delta} \boldsymbol{\alpha}\times]\big)^T+(u_i+\delta u_i)^2R_{b_i}
    \end{split}
\end{equation}
Decomposing $Q_{\hat{\lambda}_i}$ to first-order and second-order terms gives
\begin{subequations}
\begin{gather}
    Q_{\hat{\lambda}_i}=Q_{\lambda_i}+\delta Q_{\lambda_i}+\delta^2Q_{\lambda_i}\label{eqn:q_lam_i}\\
    Q_{\lambda_i}=v^2_i A R_{r_i}A^T+u^2_i R_{b_i}=S_i R_i S^T_i\label{eq:def_q_lam_i_sev}\\
    \delta Q_{\lambda_i}(\boldsymbol{\delta}\boldsymbol{\alpha})=2v_i\delta v_i A R_{r_i} A^T - v^2_i [\boldsymbol{\delta}\boldsymbol{\alpha} \times]A R_{r_i} A^T-v^2_i A R_{r_i} A^T [\boldsymbol{\delta}\boldsymbol{\alpha} \times] + 2u_i \delta u_i R_{b_i}\\
    \delta^2Q_{\lambda_i}=\delta v^2_i A R_{r_i} A^T+ v^2_i [\boldsymbol{\delta}\boldsymbol{\alpha} \times] A R_{r_i} A^T [\boldsymbol{\delta}\boldsymbol{\alpha} \times]^T- 2 v_i \delta v_i\left( [\boldsymbol{\delta}\boldsymbol{\alpha} \times] A R_{r_i} A^T+A R_{r_i} A^T [\boldsymbol{\delta}\boldsymbol{\alpha} \times]^T \right)+\delta u^2_i R_{b_i}
\end{gather}
\end{subequations}
The inverse of $Q_{\hat{\lambda}_i}$ is given by
\begin{equation}
    \label{iQ_hat}
    Q^{-1}_{\hat{\lambda_i}}=Q^{-1}_{\lambda_i}-Q^{-1}_{\lambda_i}\delta Q_{\lambda_i}Q^{-1}_{\lambda_i}-Q^{-1}_{\lambda_i}\delta^2Q_{\lambda_i}Q^{-1}_{\lambda_i}
\end{equation}
So the constant term for $Q^{-1}_{\hat{\lambda_i}}$ will be $Q^{-1}_{\lambda_i}$ for which the inverse is defined in Eq.~\eqref{eqn:q_lam_i}.
The first-order terms of estimate errors are already given in the quadratic terms in Eq.~\eqref{eqn:cost_sf_1_sev_final}. Then only a part of the $Q_{\hat{\lambda}_i}$, denoted by $Q_{\lambda_i}$, which is not a function of unknown errors $\boldsymbol{\delta}\mathbf{x}$, contributes to the second-order approximation of the cost function. The second-order cost function from the first-order estimate errors gives
\begin{equation}
    \begin{split}
        L=\frac{1}{2}\sum_{i=1}^n (\Delta u_i-\mathbf{e}_i\boldsymbol{\delta}\mathbf{x})^T R^{-1}_{u_i} (\Delta u_i-\mathbf{e}_i\boldsymbol{\delta}\mathbf{x})+ (\Delta v_i-\mathbf{f}_i\boldsymbol{\delta}\mathbf{x})^T R^{-1}_{v_i} (\Delta v_i-\mathbf{f}_i\boldsymbol{\delta}\mathbf{x})+\frac{1}{2}\sum_{i=1}^n (\boldsymbol{\Delta} \mathbf{a}_i - G_i \boldsymbol{\delta}\mathbf{x})^T Q^{-1}_{\lambda_i} (\boldsymbol{\Delta} \mathbf{a}_i - G_i \boldsymbol{\delta}\mathbf{x})\label{eq:2ndorder_sev}
    \end{split}
\end{equation}
in which 
\begin{subequations}
    \begin{gather}
        \mathbf{e}_i=\begin{bmatrix}0_{1\times (4+2i)} & 1 & 0_{1\times (2n-2i+1)}\end{bmatrix}_{1\times(6+2n)}\label{eq:def_ei}\\
    \mathbf{f}_i=\begin{bmatrix}0_{1\times (5+2i)} & 1 & 0_{1\times (2n-2i)}\end{bmatrix}_{1\times(6+2n)}
    \end{gather}
\end{subequations}
Note that Eq.~\eqref{eq:2ndorder_sev} is analogous to second-order approximation of the log-likelihood function in Eq.~\eqref{eqn:cost_TLS_D}.

\subsection{Covariance Analysis of the Estimates and Residuals}\label{cov_analysis}
In this section, the analytical expressions for the covariance of estimates and measurement residuals are derived.
The necessary conditions for the unknowns from the second-order cost function in Eq.~\eqref{eq:2ndorder_sev} yields
\begin{equation}
    \begin{split}
        \frac{\partial L}{\partial \boldsymbol{\delta} \mathbf{x}}&=-\sum_{i=1}^n \mathbf{e}^T_i R^{-1}_{u_i}\left(\Delta u_i-\mathbf{e}_i \boldsymbol{\delta}\mathbf{x}\right) -\sum_{i=1}^n \mathbf{f}^T_i R^{-1}_{v_i}\left(\Delta v_i-\mathbf{f}_i \boldsymbol{\delta}\mathbf{x}\right)-\sum_{i=1}^n G^T_i Q^{-1}_{\lambda_i}\left(\boldsymbol{\Delta}\mathbf{a}_i-G_i\boldsymbol{\delta}\mathbf{x}\right)\\
    &=F\boldsymbol{\delta}\mathbf{x}-\mathbf{g}\\
    &=\mathbf{0}_{(6+2n)\times 1}
    \end{split}
\end{equation}
Where
\begin{subequations}
\begin{gather}
    F=\sum_{i=1}^n\mathbf{e}^T_i R^{-1}_{u_i}\mathbf{e}_i+ \mathbf{f}^T_i R^{-1}_{v_i}\mathbf{f}_i+ G^T_i Q^{-1}_{\lambda_i} G_i\label{eq:def_FIM_sev}\\
    \mathbf{g}=\sum_{i=1}^n \mathbf{e}^T_i R^{-1}_{u_i} \Delta u_i+\mathbf{f}^T_i R^{-1}_{v_i} \Delta v_i+G^T_i Q^{-1}_{\lambda_i}\boldsymbol{\Delta}\mathbf{a}_i
\end{gather}
\end{subequations}
Then the optimal vector of unknowns will be
\begin{equation}
    \boldsymbol{\delta}\mathbf{x}=F^{-1}\mathbf{g}\label{eq:opt_dx}
\end{equation}
\subsubsection{Estimate Covatiance}
The estimate covariance of $\boldsymbol{\delta}\mathbf{x}$ is given 
\begin{equation}
    \begin{split}
        \text{cov}\{\boldsymbol{\delta}\mathbf{x}\}& \equiv \text{E}\{\boldsymbol{\delta}\mathbf{x}\boldsymbol{\delta}\mathbf{x}^T\}\\
    &=\text{E}\{F^{-1}\mathbf{g}\mathbf{g}^T F^{-T}\}\\
    &=F^{-1}\text{E}\{\mathbf{g}\mathbf{g}^T\}F^{-T}\label{eq:P_dx_partial_sev}
    \end{split}
\end{equation}
For calculation of $\text{E}\{\mathbf{g}\mathbf{g}^T\}$ series of identities is required, listed here:
\begin{subequations}
\begin{gather}
    \text{E}\{\boldsymbol{\Delta} \mathbf{a}_i \boldsymbol{\Delta} \mathbf{a}^T_i\}=Q_{\lambda_i}\label{eq:P_ai_sev}\\
    \text{E}\{\Delta v_i\boldsymbol{\Delta}\mathbf{a}^T_i\}=\mathbf{0}_{1\times 3}\label{eq:P_v_ai_sev}\\
    \text{E}\{\Delta u_i\boldsymbol{\Delta}\mathbf{a}^T_i\}=\mathbf{0}_{1\times 3}\label{eq:P_u_ai_sev}
\end{gather}
\end{subequations}
\textit{Proof for Eq.~\eqref{eq:P_ai_sev}}: Based on the definition of $\boldsymbol{\Delta} \mathbf{a}_i$ in Eq.~\eqref{eq:def_delta_ai_sev}
\begin{equation}
    \begin{split}
       P_{\boldsymbol{\Delta}\mathbf{a}_i}&\equiv \text{E}\{\boldsymbol{\Delta}\mathbf{a}_i\boldsymbol{\Delta}\mathbf{a}^T_i\}\\
    &=\text{E}\{\left(u_i\boldsymbol{\Delta}\mathbf{b}_i-v_i A \boldsymbol{\Delta}\mathbf{r}_i\right)\left(u_i\boldsymbol{\Delta}\mathbf{b}_i-v_i A \boldsymbol{\Delta}\mathbf{r}_i\right)^T\}\\
    &=u^2_i R_{b_i}+v^2_i A R_{r_i} A^T\\
    &=Q_{\lambda_i}\label{eq:P_delta_ai} 
    \end{split}
\end{equation}
in which the definition of $Q_{\lambda_i}$ in Eq.~\eqref{eq:def_q_lam_i_sev} and the fact that the cross-covariance $R_{rb_i}$ is zero are used. 
Eq.~\eqref{eq:P_v_ai_sev} is also based on the assumption that the cross covariance of the measured depth and the observation direction measurements $\tilde{\mathbf{r}}_i$ and $\tilde{\mathbf{b}}_i$ are zero which is denoted in Eq.~\eqref{eq:P_v_ai_sev} and Eq.~\eqref{eq:P_u_ai_sev}.
Using the derivation of estimate covariance for $\boldsymbol{\delta}\mathbf{x}$ from Eq.~\eqref{eq:P_dx_partial_sev} gives
\begin{equation}
    \begin{split}
        \text{E}\{\mathbf{g}\mathbf{g}^T\}&=\text{E}\Big\{\left(\sum_{i=1}^n \mathbf{e}^T_i R^{-1}_{u_i} \Delta u_i+\mathbf{f}^T_i R^{-1}_{v_i} \Delta v_i+G^T_i Q^{-1}_{\lambda_i}\boldsymbol{\Delta}\mathbf{a}_i\right)\left(\sum_{i=1}^n \mathbf{e}^T_i R^{-1}_{u_i} \Delta u_i+\mathbf{f}^T_i R^{-1}_{v_i} \Delta v_i+G^T_i Q^{-1}_{\lambda_i}\boldsymbol{\Delta}\mathbf{a}_i\right)^T\Big\}\\
    &=\sum_{i=1}^n \left( \mathbf{e}^T_i R^{-1}_{u_i}\text{E}\{\Delta u^2_i\}R^{-1}_{u_i} \mathbf{e}_i+\mathbf{f}^T_i R^{-1}_{v_i}\text{E}\{\Delta v^2_i\}R^{-1}_{v_i} \mathbf{f}_i \right)\\
    &+\sum_{i=1}^n G^T_i Q^{-1}_{\lambda_i} \text{E}\{\boldsymbol{\Delta}\mathbf{a}_i\Delta\mathbf{a}^T_i\}Q^{-1}_{\lambda_i}G_i\\
    &+\sum_{j=1}^n \mathbf{e}^T_j R^{-1}_{u_j}\sum_{i=1}^n \text{E}\{\Delta u_j\Delta\mathbf{a}^T_iQ^{-T}_{\lambda_i}G_i\}+\sum_{j=1}^n \mathbf{f}^T_j R^{-1}_{v_j}\sum_{i=1}^n \text{E}\{\Delta v_j\Delta\mathbf{a}^T_iQ^{-T}_{\lambda_i}G_i\}\\
    &+\Big[\sum_{j=1}^n \mathbf{e}^T_j R^{-1}_{u_j}\sum_{i=1}^n \text{E}\{\Delta u_j\Delta\mathbf{a}^T_iQ^{-T}_{\lambda_i}G_i\}+\sum_{j=1}^n \mathbf{f}^T_j R^{-1}_{v_j}\sum_{i=1}^n \text{E}\{\Delta v_j\Delta\mathbf{a}^T_iQ^{-T}_{\lambda_i}G_i\}\Big]^T\\
    &=\sum_{i=1}^n \left(\mathbf{e}^T_i R^{-1}_{u_i}R_{u_i}R^{-1}_{u_i} \mathbf{e}_i+\mathbf{f}^T_i R^{-1}_{v_i}R_{v_i}R^{-1}_{v_i} \mathbf{f}_i +G^T_i Q^{-1}_{\lambda_i} Q_{\lambda_i}Q^{-1}_{\lambda_i}G_i\right)\\
    &=\sum_{i=1}^n\left(\mathbf{e}^T_i R^{-1}_{u_i}\mathbf{e}_i+ \mathbf{f}^T_i R^{-1}_{v_i}\mathbf{f}_i+ G^T_i Q^{-1}_{\lambda_i} G_i\right)
    \end{split}
\end{equation}
in which based on Eq.~\eqref{eq:P_v_ai_sev}, the last two of the 4 terms are zero.
The fact that the correlation between the different features are zero,~i.e. $\text{E}\{\boldsymbol{\Delta} \mathbf{a}_i\Delta \mathbf{a}^T_j\}=0$ for $i\neq j$, is also used.
The remaining terms add up to match the definition of the FIM in Eq.~\eqref{eq:def_FIM_sev} and therefore from Eq.~\eqref{eq:P_dx_partial_sev}
\begin{equation}
    \begin{split}
        \text{cov}\{\boldsymbol{\delta}\mathbf{x}\}&=F^{-1}\text{E}\{\mathbf{g}\mathbf{g}^T\}F^{-T}\\
    &=F^{-1}F F^{-1}\\
    &=F^{-1}\label{eq:cov_unknown}
    \end{split}
\end{equation}

\subsubsection{Residual Covariance for Measurements}
As mentioned earlier in section , the TLS solution estimates the vector observations.
This makes the TLS a potential solution for the SLAM problem in which the location of the vehicle with respect to the environment landmarks or features is needed, in addition to solving for the vehicle's pose.
For the residual covariances of the observation vectors, Eq.~\eqref{eqn:def_meas_res} gives
\begin{equation}
    \begin{split}
        \mathbf{\tilde{d}}_i-\mathbf{\hat{d}}_i&\equiv\mathbf{d}_i+\boldsymbol{\Delta}\mathbf{d}_i-(\mathbf{d}_i+\boldsymbol{\delta}\mathbf{d}_i)\\
    &=\boldsymbol{\Delta}\mathbf{d}_i-\boldsymbol{\delta}\mathbf{d}_i\\
    &=-R_i\hat{S}^T_i Q^{-1}_{\hat{\lambda}_i}\left(\hat{u}_i\mathbf{\tilde{b}}_i-\hat{v}_i\hat{A}\mathbf{\tilde{r}}_i+\mathbf{\hat{p}}\right)\label{eq:residual_equal}
    \end{split}
\end{equation}
For the first-order approximation, Eq.~\eqref{eq:first_order_sev} yields
\begin{equation}
    \boldsymbol{\Delta}\mathbf{d}_i-\boldsymbol{\delta}\mathbf{d}_i\approx-C_i(\boldsymbol{\Delta} \mathbf{a}_i - G_i \boldsymbol{\delta}\mathbf{x})\label{eq:1st_residual_approx}
\end{equation}
where $C_i=R_i S^T_i Q^{-1}_{\lambda_i}$. Note that the terms $\hat{S}_i$ and $Q_{\hat{\lambda}_i}$ in Eq.~\eqref{eq:residual_equal} are also function of the unknown attitude and position, but their first-order approximations cannot affect the final answer for the first-order error because the first-order error terms inside of the parenthesis in Eq.~\eqref{eq:residual_equal} are already given.
Then the covariance of the measurement residual within the first order of errors will be
\begin{equation}
    \begin{split}
        \text{cov}\{\boldsymbol{\Delta}\mathbf{d}_i-\boldsymbol{\delta}\mathbf{d}_i\}&\equiv \text{E}\{(\boldsymbol{\Delta}\mathbf{d}_i-\boldsymbol{\delta}\mathbf{d}_i)(\boldsymbol{\Delta}\mathbf{d}_i-\boldsymbol{\delta}\mathbf{d}_i)^T\}\\
    &\approx C_i \text{E}\Big\{\boldsymbol{\Delta} \mathbf{a}_i - G_i \boldsymbol{\delta}\mathbf{x})(\boldsymbol{\Delta} \mathbf{a}_i - G_i \boldsymbol{\delta}\mathbf{x})^T\big\}C^T_i\\
    &=C_i \Big[ \text{E}\{\boldsymbol{\Delta} \mathbf{a}_i\boldsymbol{\Delta} \mathbf{a}^T_i\} -\text{E}\{ G_i \boldsymbol{\delta}\mathbf{x}\boldsymbol{\Delta} \mathbf{a}^T_i\}
    -\text{E}\{ G_i \boldsymbol{\delta}\mathbf{x}\boldsymbol{\Delta} \mathbf{a}^T_i\}^T+G_i \text{cov}\{\boldsymbol{\delta}\mathbf{x}\} G^T_i\Big] C^T_i
    \end{split}
\end{equation}
From Eq.~\eqref{eq:P_delta_ai} and based on the identity of
\begin{align}
    \text{E}\{G_i\boldsymbol{\delta}\mathbf{x}\boldsymbol{\Delta}\mathbf{a}^T_i\}=G_i F^{-1}G^T_i\label{eq:P_dx_Dai}
\end{align}
the residual covariance will be
\begin{equation}
    \begin{split}
        \text{cov}\{\boldsymbol{\Delta}\mathbf{d}_i-\boldsymbol{\delta}\mathbf{d}_i\}&=C_i(Q_{\lambda_i}-G_i F^{-1}G^T_i-G_i F^{-1}G^T_i+G_i F^{-1}G^T_i)C^T_i\\
    &=C_i(Q_{\lambda_i}-G_i F^{-1}G^T_i)C^T_i\label{eq:cov_di}
    \end{split}
\end{equation}
\textit{Proof for Eq.~\eqref{eq:P_dx_Dai} }: From the optimal value of $\boldsymbol{\delta}\mathbf{x}$ in Eq.~\eqref{eq:opt_dx}, the following expression is given:
\begin{equation}
    \begin{split}
        \text{E}\{G_i\boldsymbol{\delta}\mathbf{x}\boldsymbol{\Delta}\mathbf{a}^T_i\}&=\text{E}\{G_iF^{-1}\mathbf{g}\boldsymbol{\Delta}\mathbf{a}^T_i\}\\
    &=\text{E}\Big\{ G_i F^{-1} \Big[\sum_{j=1}^n \mathbf{e}^T_j R^{-1}_{u_j}\Delta u_j \boldsymbol{\Delta}\mathbf{a}^T_i+\mathbf{f}^T_j R^{-1}_{v_j}\Delta v_j \boldsymbol{\Delta}\mathbf{a}^T_i  +G^T_j Q^{-1}_{\lambda_j} \boldsymbol{\Delta}\mathbf{a}_j\boldsymbol{\Delta}\mathbf{a}^T_i \Big] \Big\}
    \end{split}
\end{equation}
From the assumption that the correlation between the depth measurement errors $\Delta u_i$ and $\Delta v_i$ and the direction measurement residuals $\boldsymbol{\Delta}\mathbf{a}_i$ are zero, also from Eq.~\eqref{eq:P_delta_ai}, the following expression is given:
\begin{equation}
    \begin{split}
       \text{E}\{G_i\boldsymbol{\delta}\mathbf{x}\boldsymbol{\Delta}\mathbf{a}^T_i\}&=\text{E}\Big\{ G_i F^{-1} \Big[\sum_{j=1}^n G^T_j Q^{-1}_{\lambda_j} \boldsymbol{\Delta}\mathbf{a}_j\boldsymbol{\Delta}\mathbf{a}^T_i \Big] \Big\}\\
    &=G_i F^{-1} G^T_i 
    \end{split}
\end{equation}
where the identity that only one term in the former summation is nonzero based on $\text{E}\{\boldsymbol{\Delta}\mathbf{a}_j\boldsymbol{\Delta}\mathbf{a}^T_i\}=0$ for $i\neq j$ is used.
This originates from the fact that different measurements from non-identical features have zero correlation with each other, which yields $\text{E}\{\boldsymbol{\Delta}\mathbf{r}_j\boldsymbol{\Delta}\mathbf{r}^T_i\}=0$ and  $\text{E}\{\boldsymbol{\Delta}\mathbf{b}_j\boldsymbol{\Delta}\mathbf{b}^T_i\}=0$ for $i \neq j$.
\subsubsection{Estimate Covariance for Measurements}
The estimate covariance for observation vectors is defined as
\begin{equation}
    P_{\boldsymbol{\delta}\mathbf{d}_i}\equiv \text{cov}\{\boldsymbol{\delta}\mathbf{d}_i\}= \text{E}\{\boldsymbol{\delta}\mathbf{d}_i\boldsymbol{\delta}\mathbf{d}^T_i\}
\end{equation}
Using the first-order approximation in Eq.~\eqref{eq:1st_residual_approx} gives
\begin{equation}
    \boldsymbol{\delta}\mathbf{d}_i\approx \boldsymbol{\Delta}\mathbf{d}_i+C_i(\boldsymbol{\Delta} \mathbf{a}_i - G_i \boldsymbol{\delta}\mathbf{x})
\end{equation}
The covariance within first order of the observation errors is
\begin{equation}
    \begin{split}
        P_{\boldsymbol{\delta}\mathbf{d}_i}&=\text{E}\Big\{\Big[\boldsymbol{\Delta}\mathbf{d}_i+C_i(\boldsymbol{\Delta} \mathbf{a}_i - G_i \boldsymbol{\delta}\mathbf{x})\Big]\Big[\boldsymbol{\Delta}\mathbf{d}_i+C_i(\boldsymbol{\Delta} \mathbf{a}_i - G_i \boldsymbol{\delta}\mathbf{x})\Big]^T\Big\}\\
    &=\text{E}\{\boldsymbol{\Delta}\mathbf{d}_i\boldsymbol{\Delta}\mathbf{d}^T_i\}+\text{cov}\{\boldsymbol{\Delta}\mathbf{d}_i-\boldsymbol{\delta}\mathbf{d}_i\}\\
    &+\text{E}\Big\{C_i(\boldsymbol{\Delta} \mathbf{a}_i - G_i \boldsymbol{\delta}\mathbf{x})\boldsymbol{\Delta\mathbf{d}^T_i}\Big\}+\text{E}\Big\{C_i(\boldsymbol{\Delta} \mathbf{a}_i - G_i \boldsymbol{\delta}\mathbf{x})\boldsymbol{\Delta\mathbf{d}^T_i}\Big\}^T
    \end{split}
\end{equation}
Then for the cross covariance terms, the following expression is given:
\begin{equation}
    \begin{split}
        \text{E}\Big\{C_i(\boldsymbol{\Delta} \mathbf{a}_i - G_i \boldsymbol{\delta}\mathbf{x})\boldsymbol{\Delta}\mathbf{d}^T_i\Big\}&=\text{E}\{C_i\boldsymbol{\Delta}\mathbf{a}_i\boldsymbol{\Delta}\mathbf{d}^T_i\}-\text{E}\{C_i G_i \boldsymbol{\delta}\mathbf{x}\boldsymbol{\Delta}\mathbf{d}^T_i\}\\
    &=\text{E}\{-C_i S_i \boldsymbol{\Delta}\mathbf{d}_i\boldsymbol{\Delta}\mathbf{d}^T_i\}-\text{E}\{C_i G_i F^{-1} \sum_{j=1}^n G^T_j Q^{-1}_{\lambda_j}\boldsymbol{\Delta}\mathbf{a}_j \boldsymbol{\Delta}\mathbf{d}^T_i\}\\
    &=-C_i S_i R_i+C_i G_i F^{-1} G^T_i C^T_i\label{eq:cross_term_P_di}
    \end{split}
\end{equation}
where the identity of $\boldsymbol{\Delta}\mathbf{a}_i=-S_i \boldsymbol{\Delta}\mathbf{d}_i$ and also the fact that only one term in the summation in Eq.~\eqref{eq:cross_term_P_di} is nonzero are used.
Then from Eq.~\eqref{eq:P_Ddi} and Eq.~\eqref{eq:cov_di}, the covariance of measurement estimate $P_{\boldsymbol{\delta}\mathbf{d}_i}$ will be
\begin{equation}
    \begin{split}
        P_{\boldsymbol{\delta}\mathbf{d}_i}&=R_i+C_i(Q_{\lambda_i}-G_i F^{-1}G^T_i)C^T_i-(C_i S_i R_i+C^T_i S^T_i R^T_i)+2C_i G_i F^{-1} G^T_i C^T_i\\
    &=R_i+C_i(Q_{\lambda_i}+G_i F^{-1}G^T_i)C^T_i-(C_i S_i R_i+R^T_i S^T_i C^T_i)\label{eq:P_di}
    \end{split}
\end{equation}

\subsection{Fisher Information Matrix}
The analytical derivations of the estimate covariances for the unknowns,~i.e. the attitude matrix $\hat{A}$, position vector $\hat{\mathbf{p}}$, depth estimates $\hat{u_i}$ and $\hat{v_i}$, as well as the observation vectors $\hat{\mathbf{r}}_i$ and $\hat{\mathbf{b}}_i$ have been provided in section \ref{cov_analysis} with the small angle approximation.
Now the optimality of the estimates using the Fisher Information Matrix (FIM) and Cram\'er-Rao Lower Bound (CRLB) \cite{cramer1999mathematical} is shown.
For an unbiased estimator $\mathbf{\hat{x}}$, the estimate error-covariance has a lower bound as
\begin{equation}
    \text{cov}\{\mathbf{\hat{x}}\}\geq F^{-1}\equiv \left(\text{E}\{-\frac{\partial^2}{\partial \mathbf{\hat{x}} \partial \mathbf{\hat{x}}^T}p(\mathbf{\tilde{y}|\mathbf{\hat{x}}})\}\right)^{-1}\label{eqn:crlb}
\end{equation}
The term inside of the expectation shows the Hessian of the the negative log-likelihood function, which is given in Eq.~\eqref{eq:2ndorder_sev}.
For an efficient estimator, the equality condition in Eq.~\eqref{eqn:crlb} should be satisfied.
The FIM is the Hessian of the loss function in Eq.~\eqref{eq:2ndorder_sev} with respect to the unknowns vector $\boldsymbol{\delta}\mathbf{x}$ and is given by the term $F$ in Eq.~\eqref{eq:def_FIM_sev}.
Note that from the derivation in Eq.~\eqref{eq:cov_unknown}, the covariance of the estimate is equal to the inverse of the FIM which shows the optimality of the estimation based on the equality condtion in Eq.~\eqref{eqn:crlb}.

\section{Sensitivity Analysis}
In this section, the sensitivity of the estimate errors and their corresponding covariances with respect to the covariance of measurement noise for virtual depths are checked.
The term \textit{virtual} is used here in a sense that the actual depth data are not available in monocular camera images. Thus a virtual depth measurement with a large covariance is instead used.
Beginning with the unknowns estimate vector $\boldsymbol{\delta} \mathbf{x}$, from Eq.~\eqref{eq:opt_dx}, gives
\begin{equation}
\begin{split}
    \frac{\partial \boldsymbol{\delta}\mathbf{x}}{\partial R_{u_i}}&=\frac{\partial F^{-1}\mathbf{g}}{\partial R_{u_i}}\\
    &=\big[\frac{\partial F^{-1}}{\partial R_{u_i}}\big]\mathbf{g}+F^{-1}\frac{\partial \mathbf{g}}{\partial R_{u_i}}\\
    &=-F^{-1}\frac{\mathbf{e}^T_i\mathbf{e}_i}{R^2_{u_i}}F^{-1}\mathbf{g}+F^{-1}\frac{\mathbf{e}^T_i \Delta u_i}{R^2_{u_i}}\\
    &=\frac{F^{-1}\mathbf{e}^T_i}{R^2_{u_i}}\left(\Delta u_i - \mathbf{e}_i \boldsymbol{\delta} \mathbf{x}\right)
\end{split}
\end{equation}
Based on the definition of $\mathbf{e}_i$ in Eq.~\eqref{eq:def_ei} yields
\begin{equation}
    \frac{\partial \boldsymbol{\delta}\mathbf{x}} {\partial R_{u_i}}=\frac{\text{cov}\{\delta u_i\}}{R^2_{u_i}}\left(\Delta u_i - \delta u_i \right)\label{eq:sens_dx_dRui}
\end{equation}
This results in a positive number for the case of $\Delta u_i > \delta u_i$, meaning that if the measurement error is larger than the estimate error, then the estimate error of unknowns grows with an increase in the virtual depth covariances.
Also it is to be noted that the virtual depth measurements are completely uncertain, i.e.~$R_{u_i}=\infty$, then the derivative in Eq.~\eqref{eq:sens_dx_dRui} is zero and the unknown error is independent of the covariance of virtual depth measurements.
Now the variations of unknowns estimate covariances with the changes in the $R_{u_i}$ can be investigated.
From Eqs.~\eqref{eq:cov_unknown} and\eqref{eq:def_ei}, the following expression is given:
\begin{equation}
    \begin{split}
        \frac{\partial \text{cov}\{\boldsymbol{\delta}\mathbf{x}\} }{\partial R_{u_i}}&=\frac{\partial F^{-1} }{\partial R_{u_i}}\\
    &=F^{-1} \frac{\mathbf{e}^T_i\mathbf{e}_i}{R^2_{u_i}} F^{-1}\\
    &=\frac{\text{cov}\{\delta u_i\}}{R^2_{u_i}}
    \end{split}
\end{equation}
which indicates that in case of a very large covariance for $u_i$, the derivative is approaching zero and the estimate covariance if not dependant on the depth measurement covariances.
This is intuitive since a depth measurement data is not provided in monocular SLAM problems.
For the covariance of the residual errors for the observation unit vectors $\mathbf{r}_i$ and $\mathbf{b}_i$, from Eq.~\eqref{eq:cov_di},  the following expression is given:
\begin{equation}
    \begin{split}
        \frac{\partial \text{cov}\{\boldsymbol{\Delta}\mathbf{d}_i-\boldsymbol{\delta}\mathbf{d}_i\}}{\partial R_{u_i}}&=C_i G_i F^{-1} \frac{\mathbf{e}^T_i\mathbf{e}_i}{R^2_{u_i}} F^{-1} G^T_i C^T_i\\
    &=C_i G_i \frac{\text{cov}\{\delta u_i\}}{R^2_{u_i}} F^{-1} G^T_i C^T_i
    \end{split}
\end{equation}
For the covariance estimates of the unit observation vectors, from Eq.~\eqref{eq:P_di},  the following expression is given:
\begin{equation}
    \begin{split}
        \frac{\partial P_{\boldsymbol{\delta}\mathbf{d}_i} }{\partial R_{u_i}}&= -C_i G_i F^{-1} \frac{\mathbf{e}^T_i\mathbf{e}_i}{R^2_{u_i}} F^{-1} G^T_i C^T_i\\
    &=-C_i G_i \frac{\text{cov}\{\delta u_i\}}{R^2_{u_i}} F^{-1} G^T_i C^T_i
    \end{split}
\end{equation}
The same analysis can be done for the covariance of virtual depth measurements in the reference frame denoted by $R_{v_i}$.
The other notable result is the singularity of the FIM in case of increasing the covariance of the depth measurements.
For this purpose, the derivative of the determinant of the FIM is calculated analytically with respect to the covariance of virtual measurement:
\begin{equation}
\begin{split}
    \frac{\partial |F|}{\partial R_{u_i}}&=|F|\text{trace}\left(F^{-1}\frac{\partial F}{\partial R_{u_i}}\right)\\
    &=|F|\text{trace}\left(\frac{-F^{-1} \mathbf{e}^T_i\mathbf{e}_i}{R^2_{u_i}}\right)\\
    &=-|F|\frac{\text{cov}\{\delta u_i\}}{R^2_{u_i}}
\end{split}
\end{equation}
Since $\text{cov}\{\delta u_i\}$ and $R_{u_i}$ are positive definite, and $|F|$ is a positive number, then the determinant is decreasing by making the virtual depth more uncertain with a larger covariance.
This means that the FIM gets closer to being singular when increasing the uncertainty in the virtual depth.
Also, the more information that is known about the depth, the better conditioned the FIM becomes.
In a real application, the virtual depths have a very large covariance and there is no certainty about them, which makes the FIM close to singular.
This results in a large $3\sigma$ bound for the position and depth errors, as will been seen in the Monte Carlo analysis in section \ref{sec_monte_carlo}. 

\section{Numerical Validation in Monte-Carlo Simulation}\label{sec_monte_carlo}

\begin{figure}
  \begin{centering}
      \subfigure[{\bf Attitude Errors}]
      {\includegraphics[width=0.49\textwidth]{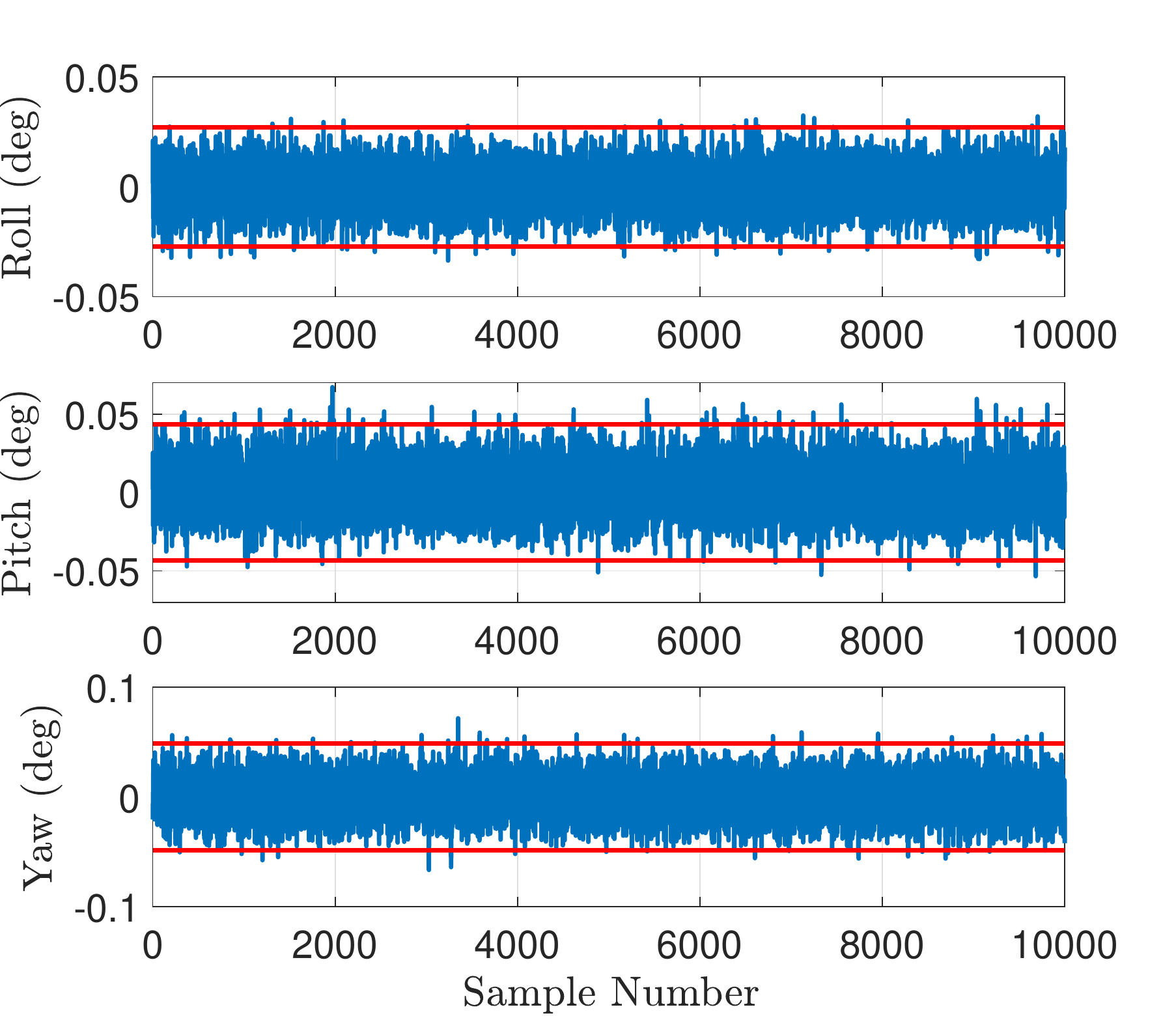}\label{fig:rpy}}
            \subfigure[{\bf Position Errors}]
      {\includegraphics[width=0.49\textwidth]{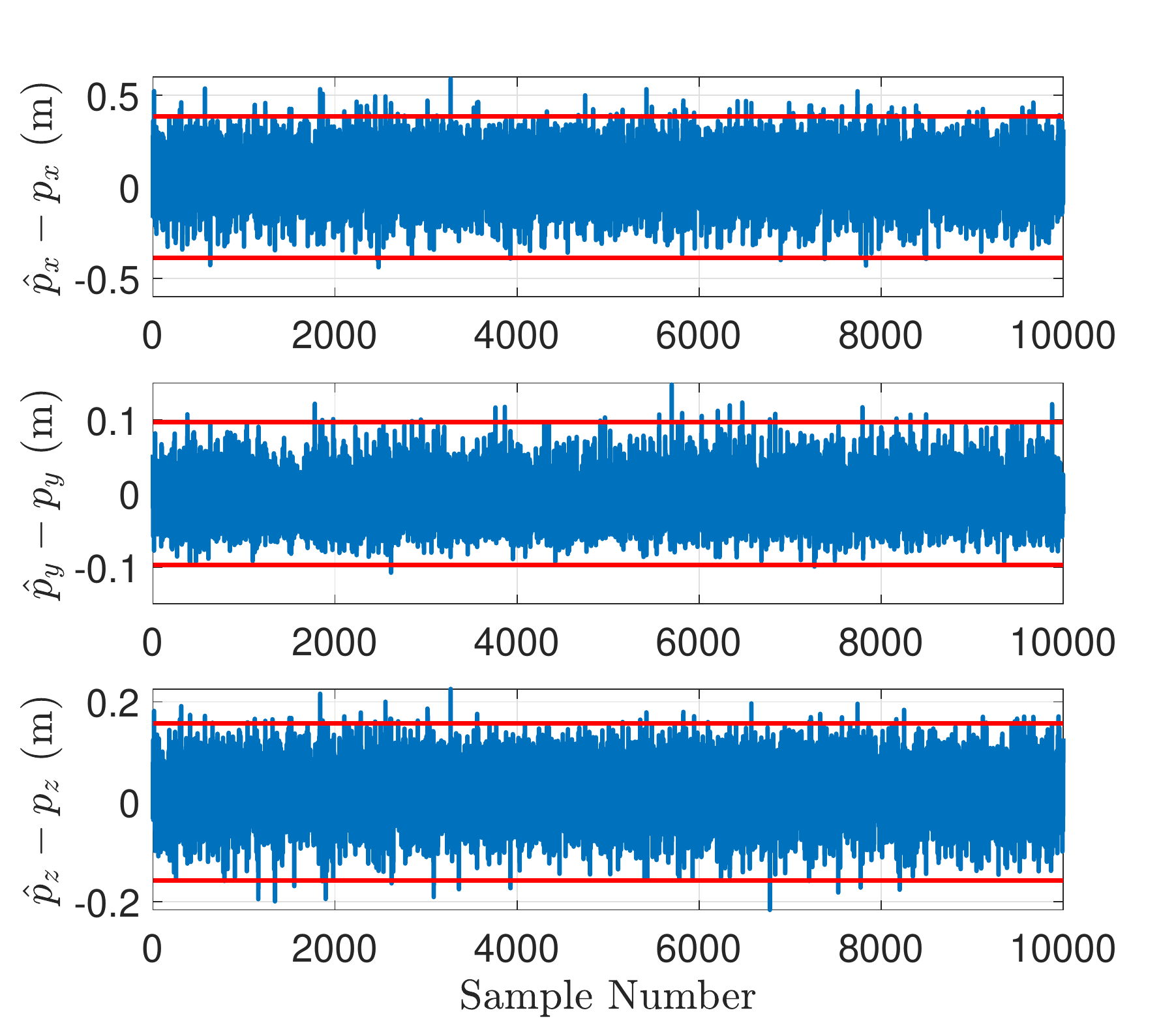}\label{fig:p}}
    \caption{Monte Carlo simulation for the attitude and translation vector.}
  \end{centering}
\end{figure}

\begin{figure}
  \begin{centering}
      \subfigure[{\bf Estimation Errors}]
      {\includegraphics[width=0.49\textwidth]{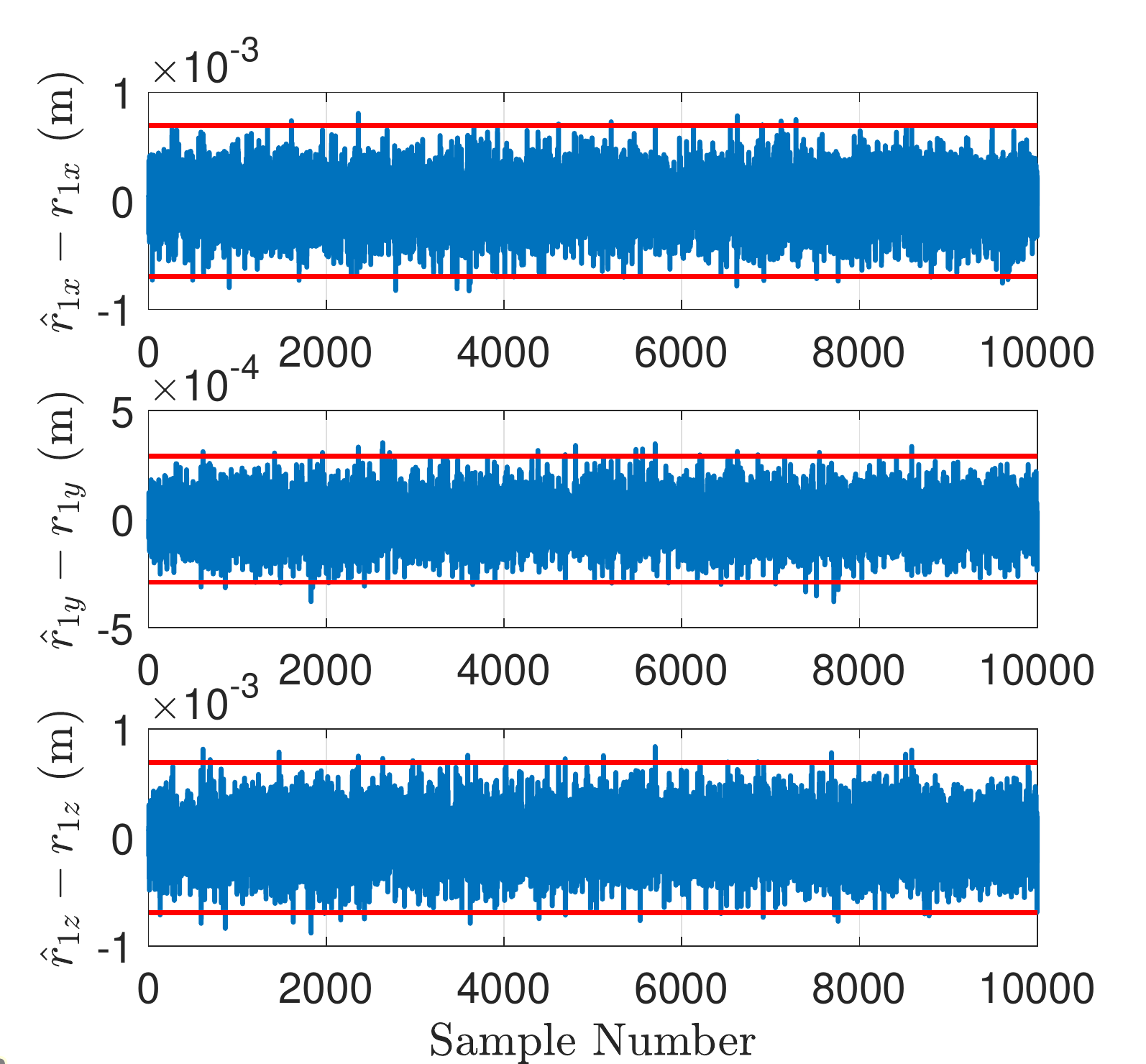}\label{fig:r_est}}
            \subfigure[{\bf Residual Errors}]
      {\includegraphics[width=0.49\textwidth]{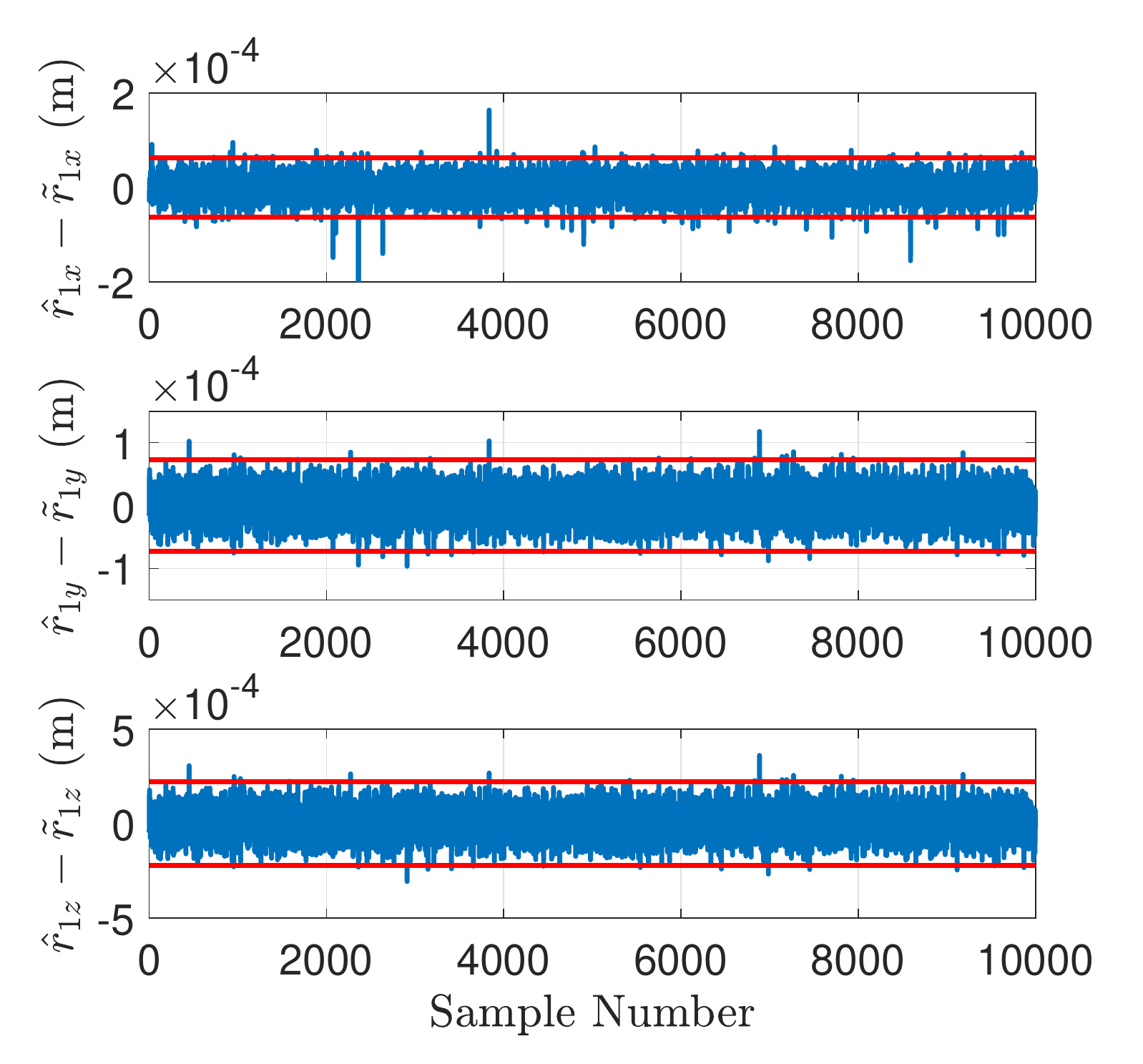}\label{fig:r_res}}
    \caption{Monte Carlo simulation for for estimates and residuals of the vector observation $\mathbf{r}_1$.}
  \end{centering}
\end{figure}

\begin{figure}
  \begin{centering}
      \subfigure[{\bf Estimation Errors}]
      {\includegraphics[width=0.49\textwidth]{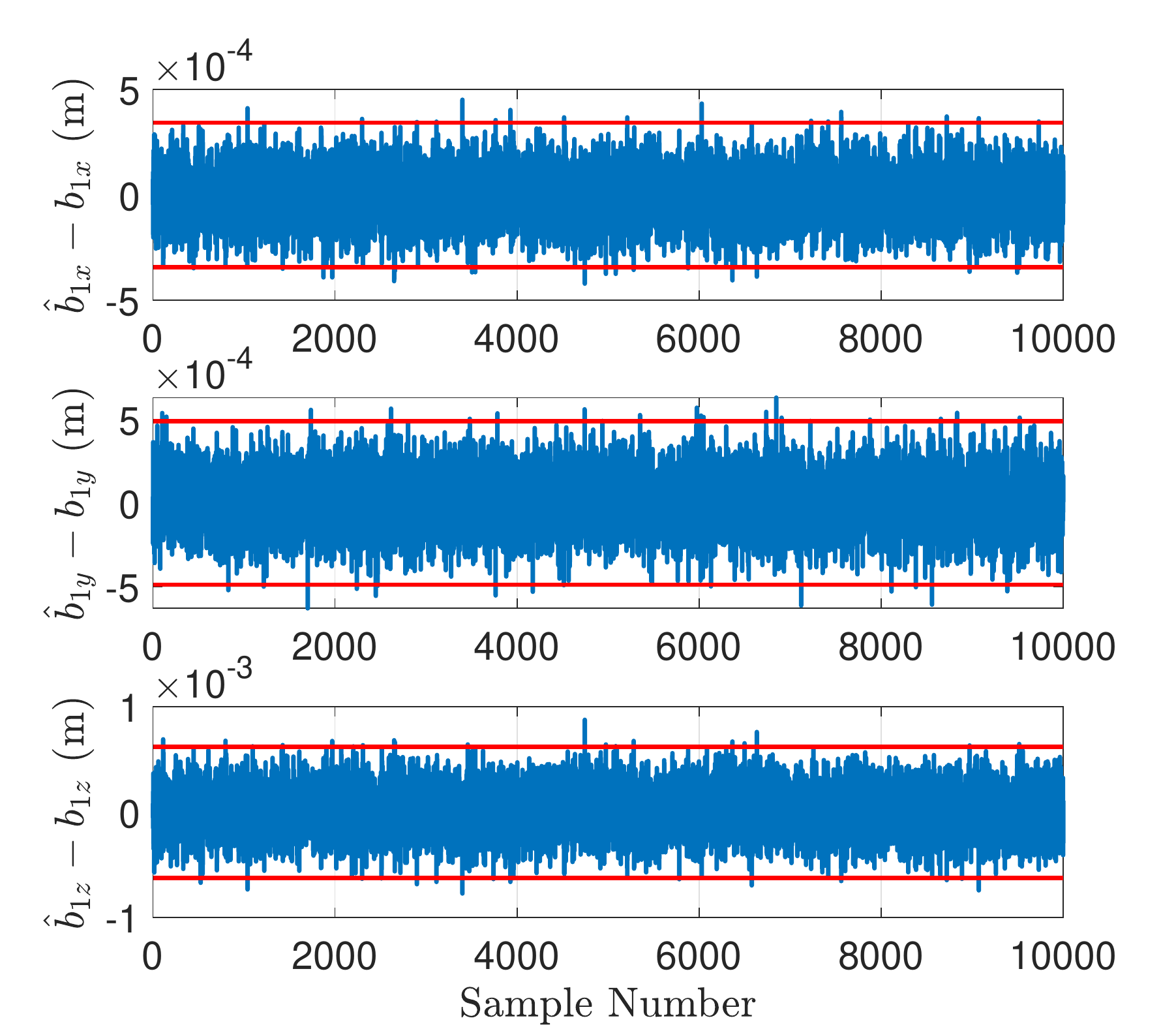}\label{fig:b_est}}
            \subfigure[{\bf Residual Errors}]
      {\includegraphics[width=0.49\textwidth]{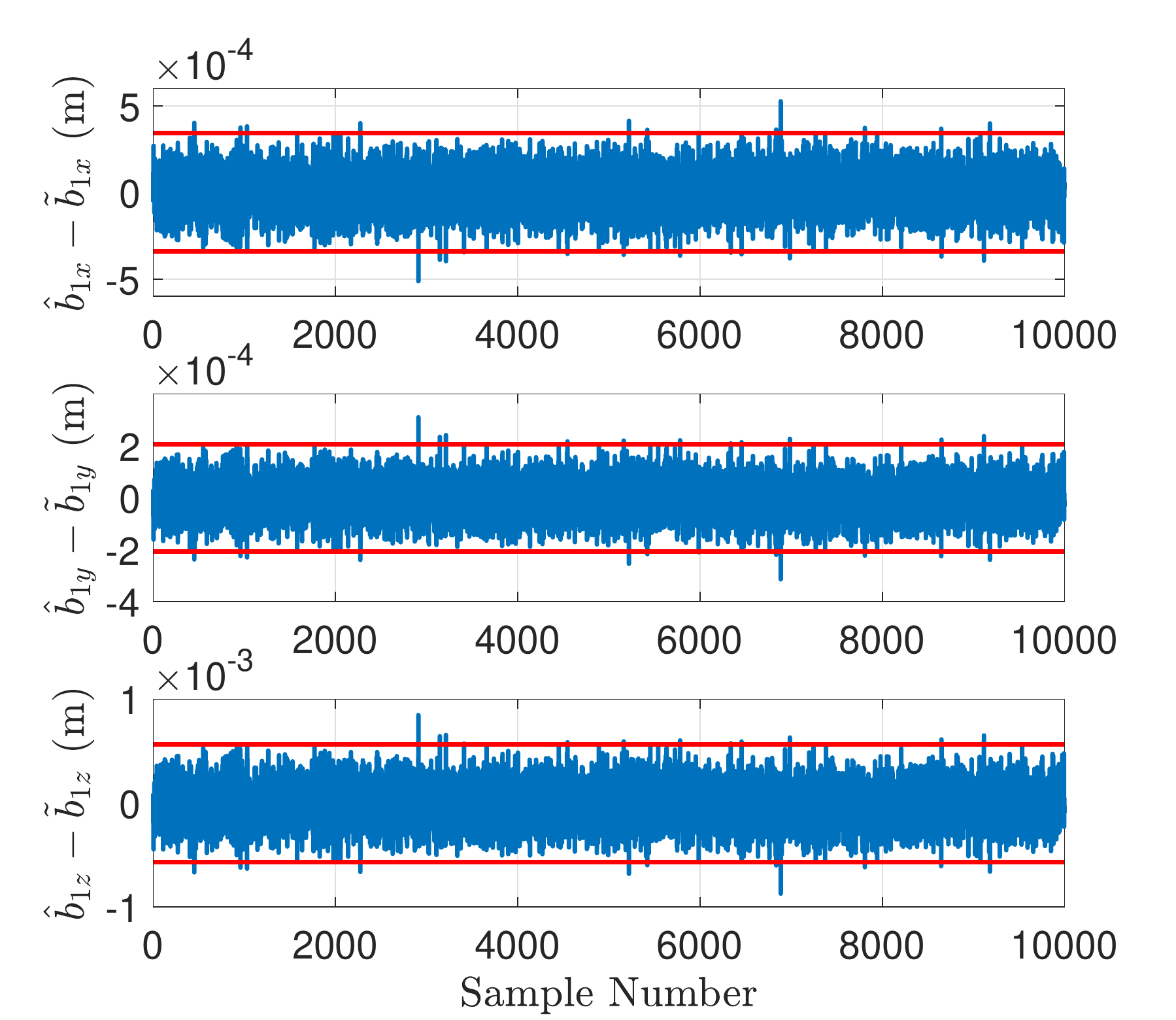}\label{fig:b_res}}
    \caption{Monte Carlo simulation for for estimates and residuals of the vector observation $\mathbf{b}_1$.}
  \end{centering}
\end{figure}

\begin{figure}
  \centering
    \includegraphics[width=\textwidth]{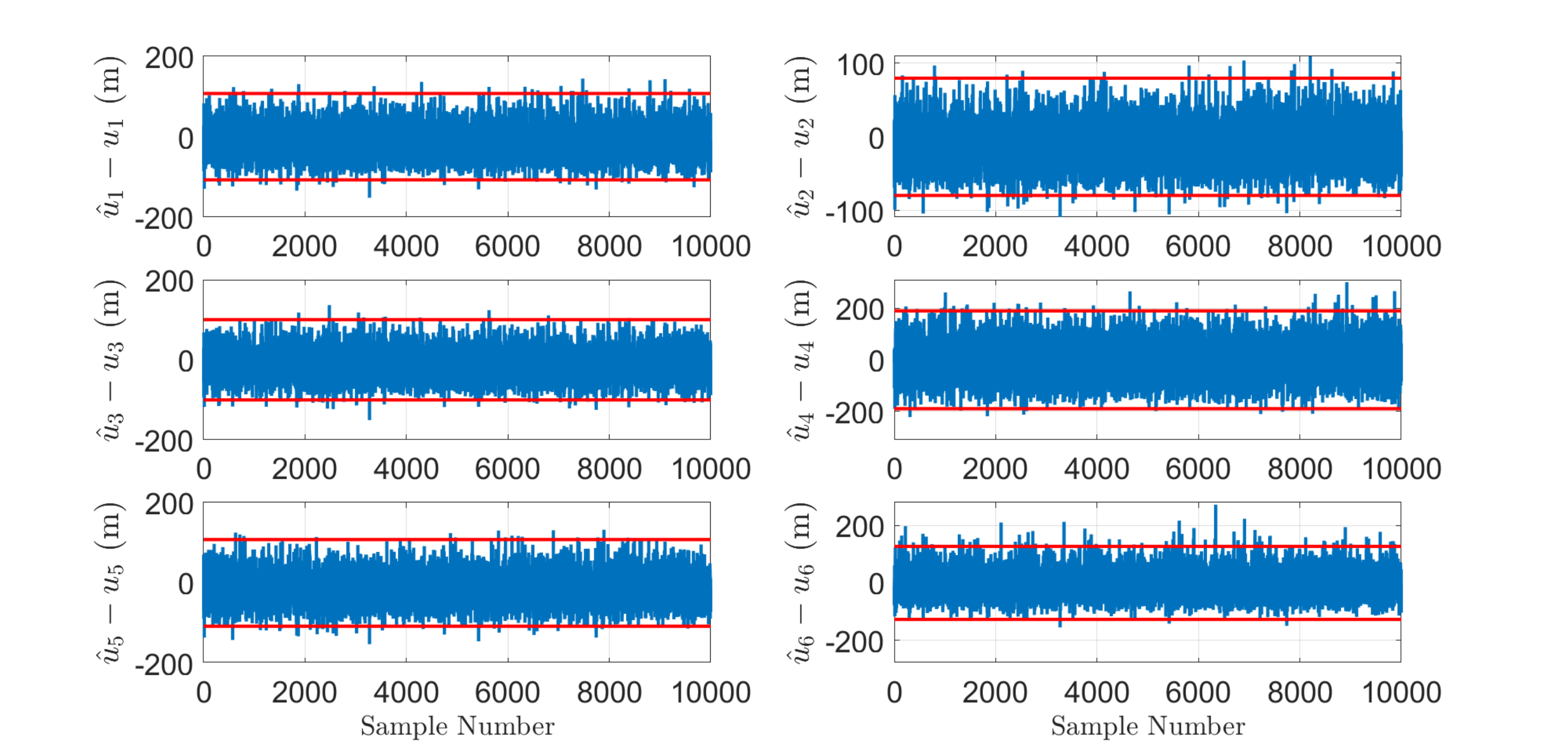}
    \caption{Monte Carlo simulation for estimates errors of virtual depth $u_i$, $i=1,...,n$.}
    \label{fig:u_k}
\end{figure}

\begin{figure}
  \begin{centering}
    \includegraphics[width=\textwidth]{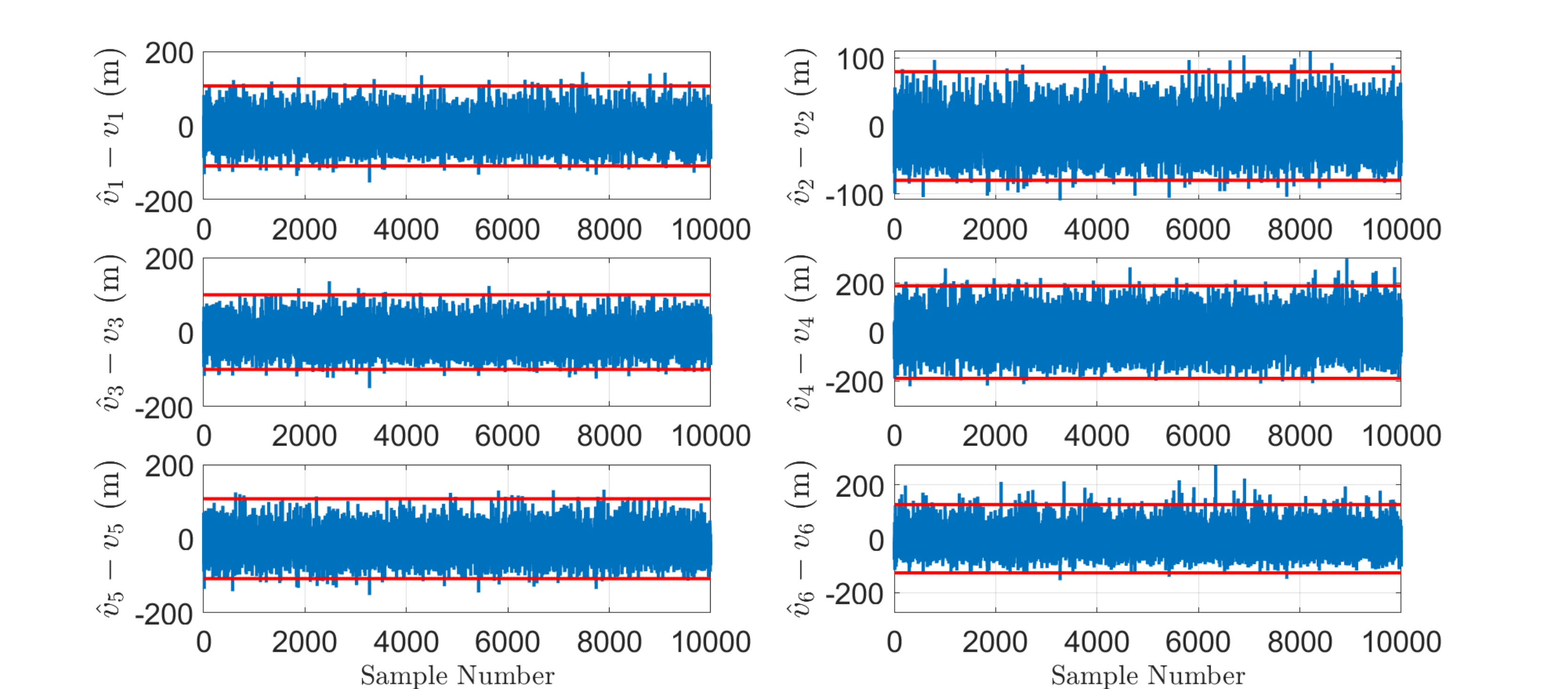}
    \caption{Monte Carlo simulation for estimates errors of virtual depth $v_i$, $i=1,...,n$.}
    \label{fig:v_k}
  \end{centering}
\end{figure}

\begin{figure}
  \begin{centering}
      {\includegraphics[width=0.6\textwidth]{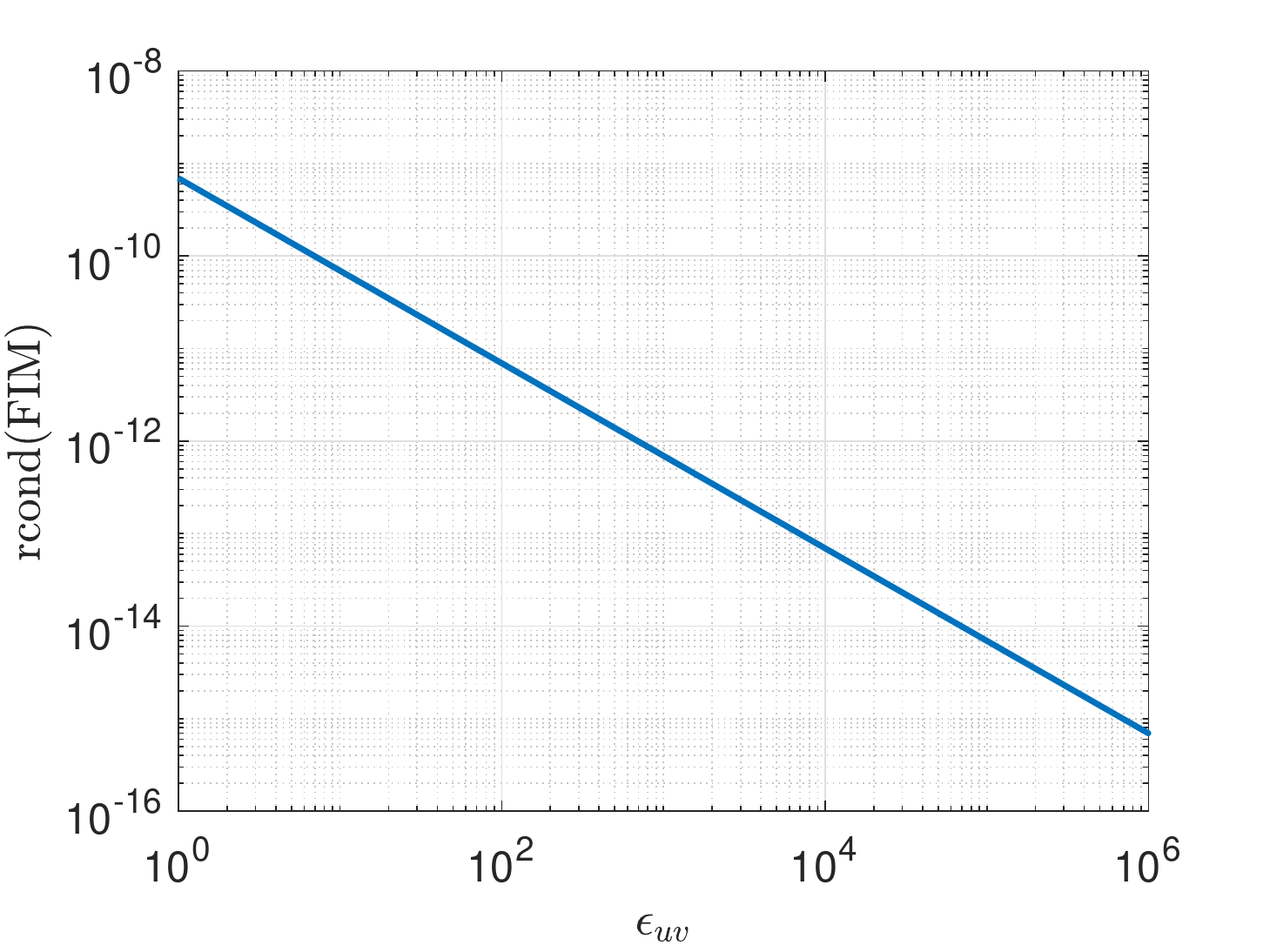}}
    \caption{The effect of virtual depth covariance on the singularity of the FIM.}
    \label{fig:cond_FIM}
  \end{centering}
\end{figure}

A pose estimation problem is solved here with two scans of a monocular camera image using the unit-sphere sensor projection model.
The minimum number of features from the simulations is three.
Using only three features, however, does not provide a good estimate in terms of the attitude error amplitudes, although the estimate errors are completely covered by the $3\sigma$ bounds from the analytical covariance derivations. In the simulations six features per image with two images are used to estimate the pose unknowns.
The ground truth values for the attitude matrix $A$, translation vector $\mathbf{p}$ and vector observations $\mathbf{b}_i$, $i=1,\,2,\,3$, are given as
    \begin{gather*}
        A=\begin{bmatrix}\cos{\frac{\pi}{4}} & \sin{\frac{\pi}{4}} & 0\\-\sin{\frac{\pi}{4}} & \cos{\frac{\pi}{4}} & 0\\0 & 0 & 1\end{bmatrix}\\
        \mathbf{p}=\begin{bmatrix}0.7512 & 1.7783 & 1.2231\end{bmatrix}^T\\
        \mathbf{r}_1=\begin{bmatrix}0.6930 & -0.0639 & 0.7181\end{bmatrix}^T\text{,\ }\mathbf{r}_2=\begin{bmatrix}0.5074 & 0.8032 & 0.3120\end{bmatrix}^T\text{,\ }\mathbf{r}_3=\begin{bmatrix}0.1558 & 0.0360 & 0.9871\end{bmatrix}^T\\
        \mathbf{r}_4=\begin{bmatrix}-0.4723 & -0.7507 & -0.4618\end{bmatrix}^T\text{,\ }\mathbf{r}_5=\begin{bmatrix}-0.9202 & -0.3649 & -0.1418\end{bmatrix}^T\text{,\ }\mathbf{r}_6=\begin{bmatrix}-0.3115 & 0.7715 & 0.5548\end{bmatrix}^T\\
        \mathbf{b}_1=\begin{bmatrix}0.3037 & -0.6373 & 0.7082\end{bmatrix}^T\text{,\ }\mathbf{b}_2=\begin{bmatrix}0.9562 & -0.0606 & 0.2863\end{bmatrix}^T\text{,\ }\mathbf{b}_3=\begin{bmatrix}0.1102 & -0.1204 & 0.9866\end{bmatrix}^T\\
        \mathbf{b}_4=\begin{bmatrix}-0.8856 & 0.0003 & -0.4645\end{bmatrix}^T\text{,\ }\mathbf{b}_5=\begin{bmatrix}-0.8020 & 0.5776 & -0.1524\end{bmatrix}^T\text{,\ }\mathbf{b}_6=\begin{bmatrix}0.4981 & 0.6662 & 0.5551\end{bmatrix}^T\\
        {u}_1=125.1371\text{,\ \ }{u}_2=35.1825\text{,\ \ }{u}_3=281.2848\text{,\ \ }{u}_4=248.2209\text{,\ \ }{u}_5=118.5592\text{,\ \ }{u}_6=67.9268\\
        {v}_1=125.1189\text{,\ \ }{v}_2=36.2025\text{,\ \ }{v}_3=282.3673\text{,\ \ }{v}_4=246.9957\text{,\ \ }{v}_5=118.8191\text{,\ \ }{v}_6=70.1661\\
    \end{gather*}

The other $u_i\text{\ , i}=1,...,n$, and $v_i\text{\ , i}=1,...,n$ values are generated by normal random vectors.
The standard deviation for the random generator of $\mathbf{r}_i\text{\ , i}=1,...,n$ is 100 meters.
The true observation vectors $\mathbf{b}_i$ are generated by the constraint in Eq.~\eqref{eqn:perfect_constraint_sev} and $u_i$'s are the norms of the projection vectors from the camera center of projection towards the feature in the environment.
A Monte-Carlo simulation with 10,000 samples is performed here to showcase how well the $3\sigma$ bounds generated by the covariance expressions in Eqs.~\eqref{eq:cov_unknown} for the attitude and position and virtual depths, cover the amplitude of the estimate errors.
Also the $3\sigma$ bounds in Eq.~\eqref{eq:cov_di} and Eq.~\eqref{eq:P_di} are checked to cover their corresponding residuals and estimate errors.
Artificial noise is generated from a Gaussian distribution with zero mean and covariance of $R_{r_i}$, $R_{b_i}$, $R_{u_i}$, $R_{v_i}$ to produce $\tilde{\mathbf{r}}_i$, $\tilde{\mathbf{b}}_i$, $\tilde{u}_i$, and $\tilde{v}_i$ samples.
The covariance matrices of the measurements $R_{r_i}$, $R_{b_i}$ are generated randomly with a coefficient of $0.006$ degrees, multiplied by a random $3\times3$ matrix, and finally multiplied by its transpose to generate a positive definite matrix.
Also the standard deviation of the depth covariances is $\epsilon_{uv}=190$ meters, which is multiplied by the absolute value of a normal Gaussian random number.
The covariances of the virtual depths are
\begin{equation}
    \begin{split}\label{eq:R_uv}
        R_{u_1}=1.5609\times 10^4 \text{,\ }R_{u_2}=1.2334\times 10^4 \text{,\ }R_{u_3}=1.2882\times 10^4\\
        R_{u_4}=9.9700\times 10^4 \text{,\ }R_{u_5}=4.8596\times 10^4 \text{,\ }R_{u_6}=1.0926\times 10^5\\
        R_{v_1}=1.9356\times 10^4 \text{,\ }R_{v_2}=6.2020\times 10^4 \text{,\ }R_{v_3}=8.1318\times 10^4\\
        R_{v_4}=3.1038\times 10^4 \text{,\ }R_{v_5}=1.1476\times 10^4 \text{,\ }R_{v_6}=4.7077\times 10^4\\
    \end{split}
\end{equation}
The covarince of line-of-sight unit vectors are
\begin{equation}
    \begin{split}
        &R_{r_1}=10^{-8}\times\begin{bmatrix}4.04& 2.53 & -0.335\\2.53 & 10.1 & -5.75\\-0.335&-5.75&4.39\end{bmatrix}\text{,\ }R_{b_1}=10^{-8}\times\begin{bmatrix}1.42& 1.44 & -1.35\\1.44 & 1.48& -1.46\\-1.35&-1.46&2.07\end{bmatrix}\\
        &R_{r_2}=10^{-8}\times\begin{bmatrix}0.15& -0.20 & 0.36\\-0.20& 4.24 & -0.08\\0.36&-0.08&3.37\end{bmatrix}\text{,\ }R_{b_2}=10^{-8}\times\begin{bmatrix}5.06& 3.03& 1.18\\3.03 & 3.22 & 1,02\\1.18&1.02&1.57\end{bmatrix}\\
        &R_{r_3}=10^{-8}\times\begin{bmatrix}8.25& 0.41 & 1.33\\0.41& 3.50 & -2.01\\1.33&-2.01&1.51\end{bmatrix}\text{,\ }R_{b_3}=10^{-8}\times\begin{bmatrix}0.83& -0.51& -0.50\\-0.51 & 0.61 & 0.67\\-0.50&0.67&5.73\end{bmatrix}\\
        &R_{r_4}=10^{-8}\times\begin{bmatrix}8.84& -2.08 & -0.06\\-2.08& 0.55 & -0.32\\-0.06&-0.32&2.10\end{bmatrix}\text{,\ }R_{b_4}=10^{-8}\times\begin{bmatrix}3.55& -1.53& 1.82\\-1.53 & 5.19 & -0.06\\-1.82&-0.06&1.17\end{bmatrix}\\
         &R_{r_5}=10^{-8}\times\begin{bmatrix}4.38& -1.93 & -3.51\\-1.93& 2.31 & -0.74\\-3.51&-0.74&6.87\end{bmatrix}\text{,\ }R_{b_5}=10^{-8}\times\begin{bmatrix}14.5& 3.18& 8.50\\3.18 & 0.80 & 1.70\\8.50&1.70&5.94\end{bmatrix}\\
          &R_{r_6}=10^{-8}\times\begin{bmatrix}7.16& 1.00 & 3.01\\1.00& 0.46 & 0.24\\3.01&0.24&1.36\end{bmatrix}\text{,\ }R_{b_6}=10^{-8}\times\begin{bmatrix}2.86& -1.04& -0.43\\-1.04 & 2.33 & -1.49\\-0.43&-1.49&1.48\end{bmatrix}
    \end{split}
\end{equation}
and then $R_{r_i}\text{\ , i}=1,...,n$ and $R_{b_i}\text{\ , i}=1,...,n$ are concatenated using Eq.~\eqref{eq:P_Ddi}.
Note that by default the measurement covariance is selected to be positive definite while being random.
Singularities in the measurement covariance matrix can be handled if they exist, but this is not the focus of this paper.

Figure \ref{fig:rpy} shows the plot of the attitude errors in terms of roll, pitch and yaw angles in degrees from the Monte-Carlo samples. The blue line depicts the estimation errors, and the red lines are the $3\sigma$ bounds computed from the estimate error-covariances. Figure \ref{fig:p} shows the translation vector estimate error in the $x$, $y$ and $z$ directions, respectively.
It is seen that the estimate errors are well-bounded by their corresponding $3\sigma$ bounds. 
Figure \ref{fig:b_est} shows the estimation errors, and Figure \ref{fig:b_res} shows the residual errors, for observation vector $\mathbf{b}_1$.
Figures \ref{fig:r_est} and \ref{fig:r_res} show the same results for the observation vector $\mathbf{r}_1$, respectively.
It is seen that the observation vectors are also bounded by their corresponding $3\sigma$ bounds, provided by the covariances of estimates as well as the residuals. 
For the virtual depth vectors, figure \ref{fig:u_k} and \ref{fig:v_k} show the estimate errors versus their corresponding analytical covariace, which shows the $3\sigma$ bounds covering the estimate errors.
Note that the estimate covariances for the virtual depths are much better than the initial covariance in Eq.~\eqref{eq:R_uv}.

The effect of the virtual depth standard deviation $\epsilon_{uv}$ on the sigularity of the FIM is shown in figure \ref{fig:cond_FIM}.
It is seen that by increasing $\epsilon_{uv}$, the reciprocal condition number of the FIM decreases, which means that the determinant is decreasing and the condition number is increasing.
The FIM is pushed more towards singularity which is intuitive, the more information that is given about the virtual depth, more accurate estimates are obtained and the FIM becomes less singular.

\section{Conclusion}
This study provides an analytical framework for an optimal estimator of monocular pose estimation, which is central to the SLAM problem.
The static SLAM problem is shown to be solved as a total least squares problem.
A quadratic cost function based on the TLS formulation is introduced for taking into account the attitude matrix, the translation vector and the virtual depth measurements with large covariance as the set of unknowns for this problem.
The weight matrix in the cost function is extracted from the most generic positive-definite fully populated matrix to include the correlations between the observation vectors in the most general case.
The cost function is then written in terms of the vector of concatenated unknowns, and the covariance expression for the attitude error is provided alongside other unknowns within the small-angle assumption and within a second-order approximation of the cost function in terms of the unknowns estimate error.
Optimal estimation of the unknowns makes the controller policy easier to track the desired signals in terms of obtaining a more accurate estimate of the states, which in turn consume lower levels of energy in the control action.
The virtual measurement are employed to avoid the observability problems.
Covariance expressions for the translation vector, position vector, virtual depths, and the estimates and residuals of the observation line-of-sight unit vectors, are obtained analytically.
The Fisher information matrix is derived, and the covariance expression of unknowns is proven to be inverse of it, which proves the equality in the Cram\'er-Rao lower bound, and thus in the optimality of the estimates.
A sensitivity analysis for the estimate errors and covariances is developed based on the perturbations in the virtual depth covariances.
This analysis shows that for the large values of the depth covariance, the estimate covariances of unknowns are independent of the depth covariance which agrees with the fact that the depth data are not available in monocular pose estimation problem. 
A simulation framework showcases the efficacy of the covariance analyses by simulating observation vectors in a pose estimation problem with 10,000 Monte-Carlo samples.
\bibliography{sample}

\begin{thebibliography}{47}
\newcommand{\enquote}[1]{``#1''}
\providecommand{\natexlab}[1]{#1}
\providecommand{\url}[1]{\texttt{#1}}
\providecommand{\urlprefix}{URL }
\expandafter\ifx\csname urlstyle\endcsname\relax
  \providecommand{\doi}[1]{\discretionary{}{}{}https://doi.org/#1}\else
  \providecommand{\doi}[1]{\discretionary{}{}{}\urlstyle{rm}\url{https://doi.org/#1}}\fi

\bibitem[{Markley and Crassidis(2014)}]{markley2014fundamentals}
Markley, F.~L., and Crassidis, J.~L., \emph{Fundamentals of Spacecraft Attitude
  Determination and Control}, Springer, 2014.

\bibitem[{Crassidis et~al.(2007)Crassidis, Markley, and
  Cheng}]{survey_nonlin_att}
Crassidis, J.~L., Markley, F.~L., and Cheng, Y., \enquote{Survey of Nonlinear
  Attitude Estimation Methods,} \emph{Journal of Guidance, Control, and
  Dynamics}, Vol.~30, No.~1, 2007, pp. 12--28.
\newblock \doi{10.2514/1.22452}.

\bibitem[{Psiaki et~al.(1990)Psiaki, Martelt, and Pal}]{psiaki1990tam}
Psiaki, M.~L., Martelt, F., and Pal, P.~K., \enquote{Three-Axis Attitude
  Determination via Kalman Filtering of Magnetometer Data,} \emph{Journal of
  Guidance, Control, and Dynamics}, Vol.~13, No.~3, 1990, pp. 506--514.
\newblock \doi{10.2514/3.25364}.

\bibitem[{Wahba(1965)}]{wahba1965least}
Wahba, G., \enquote{A Least Squares Estimate of Satellite Attitude,} \emph{SIAM
  Review}, Vol.~7, No.~3, 1965, pp. 409--409.
\newblock \doi{10.1137/1008080}.

\bibitem[{Markley and Mortari(2000)}]{markley2000quaternion}
Markley, F.~L., and Mortari, D., \enquote{Quaternion Attitude Estimation Using
  Vector Observations,} \emph{The Journal of the Astronautical Sciences},
  Vol.~48, No.~2, 2000, pp. 359--380.
\newblock \doi{10.1007/BF03546284}.

\bibitem[{Eggert et~al.(1997)Eggert, Lorusso, and
  Fisher}]{eggert1997estimating}
Eggert, D.~W., Lorusso, A., and Fisher, R.~B., \enquote{Estimating 3-D Rigid
  Body Transformations: A Comparison of Four Major Algorithms,} \emph{Machine
  Vision and Applications}, Vol.~9, No.~5, 1997, pp. 272--290.
\newblock \doi{10.1007/s001380050048}.

\bibitem[{Kelsey et~al.(2006)Kelsey, Byrne, Cosgrove, Seereeram, and
  Mehra}]{kelsey2006vision}
Kelsey, J.~M., Byrne, J., Cosgrove, M., Seereeram, S., and Mehra, R.~K.,
  \enquote{Vision-Based Relative Pose Estimation for Autonomous Rendezvous and
  Docking,} \emph{2006 IEEE Aerospace Conference}, IEEE, 2006.
\newblock \doi{10.1109/AERO.2006.1655916}.

\bibitem[{Jurie et~al.(2002)Jurie, Dhome et~al.}]{jurie2002real}
Jurie, F., Dhome, M., et~al., \enquote{Real Time Robust Template Matching,}
  \emph{BMVC}, Vol. 2002, 2002, pp. 123--132.
\newblock \doi{10.5244/C.16.10}.

\bibitem[{Isard and Blake(1998)}]{isard1998condensation}
Isard, M., and Blake, A., \enquote{Condensation—Conditional Density
  Propagation for Visual Tracking,} \emph{International Journal of Computer
  Vision}, Vol.~29, No.~1, 1998, pp. 5--28.
\newblock \doi{10.1023/A:1008078328650}.

\bibitem[{Schonberger and Frahm(2016)}]{schonberger2016structure}
Schonberger, J.~L., and Frahm, J.-M., \enquote{Structure-From-Motion
  Revisited,} \emph{Proceedings of the IEEE Conference on Computer Vision and
  Pattern Recognition}, 2016, pp. 4104--4113.
\newblock \doi{10.1109/CVPR.2016.445}.

\bibitem[{Rehbinder and Ghosh(2003)}]{rehbinder2003pose}
Rehbinder, H., and Ghosh, B.~K., \enquote{Pose Estimation Using Line-based
  Dynamic Vision and Inertial Sensors,} \emph{IEEE Transactions on Automatic
  Control}, Vol.~48, No.~2, 2003, pp. 186--199.
\newblock \doi{10.1109/TAC.2002.808464}.

\bibitem[{Sharma and D'Amico(2016)}]{sharma2016comparative}
Sharma, S., and D'Amico, S., \enquote{Comparative Assessment of Techniques for
  Initial Pose Estimation Using Monocular Vision,} \emph{Acta Astronautica},
  Vol. 123, 2016, pp. 435--445.
\newblock \doi{10.1016/j.actaastro.2015.12.032}.

\bibitem[{Lasenby et~al.(1998)Lasenby, Fitzgerald, Lasenby, and
  Doran}]{lasenby1998new}
Lasenby, J., Fitzgerald, W.~J., Lasenby, A.~N., and Doran, C., \enquote{New
  Geometric Methods for Computer Vision: An Application to Structure and Motion
  Estimation,} \emph{International Journal of Computer Vision}, Vol.~26, No.~3,
  1998, pp. 191--213.
\newblock \doi{10.1023/A:1007901028047}.

\bibitem[{Huang and Netravali(2002)}]{huang2002motion}
Huang, T.~S., and Netravali, A.~N., \enquote{Motion and Structure From Feature
  Correspondences: A Review,} \emph{Advances In Image Processing And
  Understanding: A Festschrift for Thomas S Huang}, 2002, pp. 331--347.
\newblock \doi{10.1142/9789812776952_0013}.

\bibitem[{Fu et~al.(2017)Fu, Quan, and Cai}]{fu2017robust}
Fu, Q., Quan, Q., and Cai, K.-Y., \enquote{Robust Pose Estimation for
  Multirotor UAVs Using Off-board Monocular Vision,} \emph{IEEE Transactions on
  Industrial Electronics}, Vol.~64, No.~10, 2017, pp. 7942--7951.
\newblock \doi{10.1109/TIE.2017.2696482}.

\bibitem[{Melekhov et~al.(2017)Melekhov, Ylioinas, Kannala, and
  Rahtu}]{melekhov2017relative}
Melekhov, I., Ylioinas, J., Kannala, J., and Rahtu, E., \enquote{Relative
  Camera Pose Estimation Using Convolutional Neural Networks,}
  \emph{International Conference on Advanced Concepts for Intelligent Vision
  Systems}, Springer, 2017, pp. 675--687.
\newblock \doi{10.1007/978-3-319-70353-4_57}.

\bibitem[{Li et~al.(2018)Li, Wang, Ji, Xiang, and Fox}]{li2018deepim}
Li, Y., Wang, G., Ji, X., Xiang, Y., and Fox, D., \enquote{Deepim: Deep
  Iterative Matching for 6d Pose Estimation,} \emph{Proceedings of the European
  Conference on Computer Vision (ECCV)}, 2018, pp. 683--698.
\newblock \doi{10.1007/s11263-019-01250-9}.

\bibitem[{Toshev and Szegedy(2014)}]{toshev2014deeppose}
Toshev, A., and Szegedy, C., \enquote{Deeppose: Human Pose Estimation via Deep
  Neural Networks,} \emph{Proceedings of the IEEE Conference on Computer Vision
  and Pattern Recognition}, 2014, pp. 1653--1660.
\newblock \doi{10.1109/CVPR.2014.214}.

\bibitem[{Zhang et~al.(2010)Zhang, Cao, Zhang, and He}]{zhang2010monocular}
Zhang, S., Cao, X., Zhang, F., and He, L., \enquote{Monocular Vision-based
  Iterative Pose Estimation Algorithm From Corresponding Feature Points,}
  \emph{Science China Information Sciences}, Vol.~53, No.~8, 2010, pp.
  1682--1696.
\newblock \doi{10.1007/s11432-010-4017-6}.

\bibitem[{Dornaika and Garcia(1999)}]{dornaika1999pose}
Dornaika, F., and Garcia, C., \enquote{Pose Estimation Using Point and Line
  Correspondences,} \emph{Real-Time Imaging}, Vol.~5, No.~3, 1999, pp.
  215--230.
\newblock \doi{10.1006/rtim.1997.0117}.

\bibitem[{Erol et~al.(2007)Erol, Bebis, Nicolescu, Boyle, and
  Twombly}]{ErolAli2007Vhpe}
Erol, A., Bebis, G., Nicolescu, M., Boyle, R.~D., and Twombly, X.,
  \enquote{Vision-Based Hand Pose Estimation: A Review,} \emph{Computer Vision
  and Image Understanding}, Vol. 108, No.~1, 2007, pp. 52--73.
\newblock \doi{10.1016/j.cviu.2006.10.012}.

\bibitem[{Forsyth and Ponce(2012)}]{forsyth2012computer}
Forsyth, D., and Ponce, J., \emph{Computer Vision: A Modern Approach}, Vol.~2,
  2012.

\bibitem[{Dhanachandra et~al.(2015)Dhanachandra, Manglem, and
  Chanu}]{dhanachandra2015image}
Dhanachandra, N., Manglem, K., and Chanu, Y.~J., \enquote{Image Segmentation
  Using K-means Clustering Algorithm and Subtractive Clustering Algorithm,}
  \emph{Procedia Computer Science}, Vol.~54, 2015, pp. 764--771.
\newblock \doi{10.1016/j.procs.2015.06.090}.

\bibitem[{Chen et~al.(2017)Chen, Papandreou, Schroff, and
  Adam}]{chen2017rethinking}
Chen, L.-C., Papandreou, G., Schroff, F., and Adam, H., \enquote{Rethinking
  Atrous Convolution for Semantic Image Segmentation,} \emph{arXiv preprint
  arXiv:1706.05587}, 2017.

\bibitem[{Plath et~al.(2009)Plath, Toussaint, and Nakajima}]{plath2009multi}
Plath, N., Toussaint, M., and Nakajima, S., \enquote{Multi-Class Image
  Segmentation Using Conditional Random Fields and Global Classification,}
  \emph{Proceedings of the 26th Annual International Conference on Machine
  Learning}, 2009, pp. 817--824.
\newblock \doi{10.1145/1553374.1553479}.

\bibitem[{Minaee et~al.(2021)Minaee, Boykov, Porikli, Plaza, Kehtarnavaz, and
  Terzopoulos}]{minaee2021image}
Minaee, S., Boykov, Y.~Y., Porikli, F., Plaza, A.~J., Kehtarnavaz, N., and
  Terzopoulos, D., \enquote{Image Segmentation Using Deep Learning: A Survey,}
  \emph{IEEE Transactions on Pattern Analysis and Machine Intelligence}, 2021.
\newblock \doi{0.1109/TPAMI.2021.3059968}.

\bibitem[{Chouhan et~al.(2018)Chouhan, Kaul, and Singh}]{chouhan2018soft}
Chouhan, S.~S., Kaul, A., and Singh, U.~P., \enquote{Soft Computing Approaches
  for Image Segmentation: A survey,} \emph{Multimedia Tools and Applications},
  Vol.~77, No.~21, 2018, pp. 28483--28537.
\newblock \doi{10.1007/s11042-018-6005-6}.

\bibitem[{Zaitoun and Aqel(2015)}]{zaitoun2015survey}
Zaitoun, N.~M., and Aqel, M.~J., \enquote{Survey on Image Segmentation
  Techniques,} \emph{Procedia Computer Science}, Vol.~65, 2015, pp. 797--806.
\newblock \doi{10.1016/j.procs.2015.09.027}.

\bibitem[{Canny(1986)}]{canny1986computational}
Canny, J., \enquote{A Computational Approach to Edge Detection,} \emph{IEEE
  Transactions on Pattern Analysis and Machine Intelligence}, Vol. PAMI-8,
  No.~6, 1986, pp. 679--698.
\newblock \doi{10.1109/TPAMI.1986.4767851}.

\bibitem[{Lowe(2004)}]{lowe2004distinctive}
Lowe, D.~G., \enquote{Distinctive Image Features from Scale-invariant
  Keypoints,} \emph{International Journal of Computer Vision}, Vol.~60, No.~2,
  2004, pp. 91--110.
\newblock \doi{10.1023/B:VISI.0000029664.99615.94}.

\bibitem[{Bay et~al.(2006)Bay, Tuytelaars, and Van~Gool}]{bay2006surf}
Bay, H., Tuytelaars, T., and Van~Gool, L., \enquote{Surf: Speeded Up Robust
  Features,} \emph{European Conference on Computer Vision}, Springer, 2006, pp.
  404--417.
\newblock \doi{10.1007/11744023_32}.

\bibitem[{Hough()}]{hough1962method}
Hough, P.~V., \enquote{Method and Means for Recognizing Complex Patterns,} ,
  ????
\newblock US Patent 3,069,654, Dec.~18, 1962.

\bibitem[{Ballard(1981)}]{ballard1981generalizing}
Ballard, D.~H., \enquote{Generalizing the Hough Transform to Detect Arbitrary
  Shapes,} \emph{Pattern Recognition}, Vol.~13, No.~2, 1981, pp. 111--122.
\newblock \doi{10.1016/0031-3203(81)90009-1}.

\bibitem[{Salahat and Qasaimeh(2017)}]{salahat2017recent}
Salahat, E., and Qasaimeh, M., \enquote{Recent Advances in Features Extraction
  and Description Algorithms: A Comprehensive Survey,} \emph{2017 IEEE
  International Conference on Industrial Technology (ICIT)}, IEEE, 2017, pp.
  1059--1063.
\newblock \doi{10.1109/ICIT.2017.7915508}.

\bibitem[{Davison et~al.(2007)Davison, Reid, Molton, and
  Stasse}]{davison2007monoslam}
Davison, A.~J., Reid, I.~D., Molton, N.~D., and Stasse, O., \enquote{MonoSLAM:
  Real-time Single Camera SLAM,} \emph{IEEE Transactions on Pattern Analysis
  and Machine Intelligence}, Vol.~29, No.~6, 2007, pp. 1052--1067.
\newblock \doi{10.1109/TPAMI.2007.1049}.

\bibitem[{Bian et~al.(2017)Bian, Lin, Matsushita, Yeung, Nguyen, and
  Cheng}]{Bian_2017_CVPR}
Bian, J., Lin, W.-Y., Matsushita, Y., Yeung, S.-K., Nguyen, T.-D., and Cheng,
  M.-M., \enquote{GMS: Grid-based Motion Statistics for Fast, Ultra-Robust
  Feature Correspondence,} \emph{Proceedings of the IEEE Conference on Computer
  Vision and Pattern Recognition (CVPR)}, 2017.
\newblock \doi{10.1007/s11263-019-01280-3}.

\bibitem[{Zhao et~al.(2019)Zhao, Yang, Xiao, and Cao}]{zhao2019comparative}
Zhao, C., Yang, J., Xiao, Y., and Cao, Z., \enquote{Comparative Evaluation of
  2D Feature Correspondence Selection Algorithms,} \emph{arXiv preprint
  arXiv:1904.13383}, 2019.

\bibitem[{Hashim(2020)}]{hashim2020attitude}
Hashim, H.~A., \enquote{Attitude Determination and Estimation Using Vector
  Observations: Review, Challenges and Comparative Results,} \emph{arXiv
  preprint arXiv:2001.03787}, 2020.

\bibitem[{Crassidis and Junkins(2012)}]{alma9938879548204803}
Crassidis, J.~L., and Junkins, J.~L., \emph{Optimal Estimation of Dynamic
  Systems}, 2\textsuperscript{nd} ed., Chapman and Hall/CRC, Boca Raton, Fl,
  2012, Chap.~2.
\newblock \doi{10.1201/b11154}.

\bibitem[{Maleki et~al.(2022)Maleki, Crassidis, Cheng, and
  Schmid}]{maleki2021total}
Maleki, S., Crassidis, J.~L., Cheng, Y., and Schmid, M., \enquote{Total Least
  Squares for Optimal Pose Estimation,} \emph{AIAA Scitech 2022 Forum}, AIAA,
  Reston, VA, 2022.
\newblock \doi{10.2514/6.2022-1222}.

\bibitem[{Maleki(2022)}]{maleki2022thesis}
Maleki, S., \enquote{Thesis: Total Least Squares for Optimal Pose Estimation,}
  \emph{ProQuest Dissertations Publishing}, 2022.

\bibitem[{Cheng and Crassidis(2021)}]{cheng2021optimal}
Cheng, Y., and Crassidis, J.~L., \enquote{Optimal Pose Estimation with
  Error-Covariance Analysis,} \emph{AIAA Scitech 2021 Forum}, 2021.
\newblock \doi{10.2514/6.2021-1758}.

\bibitem[{Crassidis and Cheng(2019)}]{crassidis2019maximum}
Crassidis, J.~L., and Cheng, Y., \enquote{Maximum Likelihood Analysis of the
  Total Least Squares Problem with Correlated Errors,} \emph{Journal of
  Guidance, Control, and Dynamics}, Vol.~42, No.~6, 2019, pp. 1204--1217.
\newblock \doi{10.2514/6.2019-1931}.

\bibitem[{Golub(1973)}]{golub1973some}
Golub, G.~H., \enquote{Some Modified Matrix Eigenvalue Problems,} \emph{SIAM
  Review}, Vol.~15, No.~2, 1973, pp. 318--334.
\newblock \doi{10.1137/1015032}.

\bibitem[{Markovsky and Van~Huffel(2007)}]{markovsky2007overview}
Markovsky, I., and Van~Huffel, S., \enquote{Overview of Total Least-Squares
  Methods,} \emph{Signal Processing}, Vol.~87, No.~10, 2007, pp. 2283--2302.
\newblock \doi{10.1016/j.sigpro.2007.04.004}.

\bibitem[{Golub and Van~Loan(1980)}]{golub1980analysis}
Golub, G.~H., and Van~Loan, C.~F., \enquote{An Analysis of the Total Least
  Squares Problem,} \emph{SIAM Journal on Numerical Analysis}, Vol.~17, No.~6,
  1980, pp. 883--893.
\newblock \doi{10.1137/0717073}.

\bibitem[{Cram{\'e}r(1999)}]{cramer1999mathematical}
Cram{\'e}r, H., \emph{Mathematical Methods of Statistics}, Vol.~43, Princeton
  University Press, 1999.
\newblock \doi{10.1515/9781400883868}.

\end{thebibliography}

\end{document}